\pdfoutput=1
\documentclass{article}
\usepackage{iclr2027_conference}
\usepackage[T1]{fontenc}

\usepackage[utf8]{inputenc}
\usepackage[T1]{fontenc}
\usepackage{microtype}
\usepackage{amsmath,amssymb,amsfonts}
\usepackage{booktabs}
\usepackage{graphicx}
\usepackage{float}
\usepackage{multirow}
\usepackage{tabularx}
\usepackage{array}
\usepackage{xcolor}
\usepackage{adjustbox}
\usepackage{algorithm,algorithmic}
\usepackage{placeins}
\usepackage{listings}
\usepackage[hidelinks]{hyperref}
\usepackage{url}

\newcommand{\ECPO}{\textsc{ECPO}}
\newcommand{\SlotEvid}{\textsc{Slot-EVIDCONS}}
\newcommand{\vCert}{\textsc{vCertNDCG}}
\newcommand{\SetRecover}{\textsc{SetRecover}}
\newcommand{\RankRecover}{\textsc{RankRecover}}
\newcommand{\FeasibleRate}{\textsc{FeasibleRate}}
\newcommand{\AbstainRate}{\textsc{AbstainRate}}
\newcommand{\ReExtractFaithfulness}{\textsc{ReExtractFaithfulness}}
\newcolumntype{Y}{>{\raggedright\arraybackslash}X}
\newcommand{\best}[1]{\textbf{#1}}
\newcommand{\second}[1]{\underline{#1}}
\newcommand{\artifact}[1]{\nolinkurl{#1}}
\renewcommand{\arraystretch}{1.04}
\title{\ECPO: Evidence-Coupled Policy Optimization for Verifier-Audited Candidate Ranking}

\author{Miaobo Hu$^{1,2,*}$, Shuhao Hu$^{2}$, BoKun Wang$^{2}$, Yina Sa$^{2}$,\\
Xin Wang$^{2}$, Xiaobo Guo$^{2}$, Daren Zha$^{2}$\\[4pt]
{\normalfont $^{1}$School of Artificial Intelligence, University of Chinese Academy of Sciences, Beijing, China}\\
{\normalfont $^{2}$Institute of Information Engineering, Chinese Academy of Sciences, Beijing, China}\\
{\normalfont $^{*}$Corresponding author: \texttt{humiaobo@iie.ac.cn}}}

\iclrfinalcopy
\hypersetup{pdfauthor={Miaobo Hu, Shuhao Hu, BoKun Wang, Yina Sa, Xin Wang, Xiaobo Guo, Daren Zha}}
\begin{document}
\maketitle
\lhead{Preprint}

\begin{abstract}
Ranking systems used for decision support should expose evidence that can be audited independently of the model's ranking scores. We study a closed-roster setting in which a policy jointly emits a Top-$K$ ranking and candidate-linked, span-grounded evidence bundles. Evidence-Coupled Policy Optimization (\ECPO) combines listwise utility, certificate validity, and an evidence-cycle reward that removes answer-bearing fields before deterministic recovery. On MAVEN-ERE and RAMS, \ECPO{} improves verifier-audited traceability, rank-blind recovery, and human comparative support relative to matched controls in the tested provenance settings. The gains persist under held-out evaluators, other 8B/14B backbones, and a HotpotQA answer-ranking interface. Ordinary NDCG and runtime remain explicit trade-offs.
\end{abstract}

\section{Introduction}
\label{sec:intro}

Ranking is often deployed where a score alone is insufficient.  An analyst may need to know which source spans support a candidate, whether those spans belong to the exposed evidence scope, and whether the evidence distinguishes that candidate from alternatives.  A fluent post-hoc explanation does not provide this contract: plausibility, attribution, and faithfulness can diverge~\citep{ribeiro_should_2016,jacovi_faithfully_2020,maynez_faithfulness_2020,faithtests2023}.  We therefore study a stricter output interface in which a policy jointly produces an ordered candidate list and machine-checkable evidence bundles.

Existing preference-optimization methods mainly optimize scalar preference signals and therefore assume that reward-model supervision is reliable.  When outputs contain unverifiable claims or fabricated evidence, a high scalar score can reinforce unsupported behavior.  ECPO treats verification consistency as part of the optimization target rather than a post-hoc filtering step.

At ranking time there is a finite window-local roster, every candidate has candidate-linked provenance units, and a declared support contract specifies what may count as evidence after direct candidate claims are removed.  Our primary instantiations use frozen event extraction over MAVEN-ERE~\citep{wang_maven-ere_2022}, which extends MAVEN~\citep{wang_maven_2020}, and RAMS~\citep{ebner_multi-sentence_2020}: a predefined skeleton supplies step and role semantics, while trigger and argument offsets provide provenance.  These define the evaluated interface.

The central ambiguity is what an evidence metric actually establishes.  A bundle can be parseable but unsupported, traceable to candidate $A$ but uninformative about why $A$ outranks $B$, or sufficient for a human while failing a particular automatic matcher.  We consequently separate five properties: \emph{validity} under the output schema; \emph{traceability} to a candidate trajectory; \emph{slot recovery} when a bundle remains at its submitted position; \emph{rank-blind comparative recovery} after positions and candidate--bundle pairing are removed; and \emph{human support}.  We call this slot-wise metric \SlotEvid{}@10 and reserve stronger language for the rank-blind and human audits.

\ECPO{} treats the policy action as the joint object $(\text{ranking},\text{certificate})$.  A skeleton-conditioned reward model scores candidate trajectories.  A deterministic validator checks schema and provenance.  Finally, an evidence-cycle verifier strips candidate claims, generated event IDs, ranks, and policy scores, then scores each remaining bundle against the full roster.  Group-normalized policy optimization rewards utility, certificate validity, and recovery of the selected candidate.  The gradient estimator, dynamic program, Bradley--Terry term, and KL regularization are established components; our contribution is the task interface, evidence-coupled objective, and audit protocol.

Three empirical findings summarize the evaluation.  First, \ECPO{} improves verifier-masked ranking and rank-blind recovery over GRPO and a strong reward-model ranker with deterministic evidence optimization.  Second, the advantage persists under held-out deterministic and raw-text verifiers, cross-backbone tests, a HotpotQA provenance schema, counterfactual evidence, and a blinded human audit.  Third, the deterministic pipeline remains slightly stronger on ordinary NDCG and far cheaper.  Together, these evaluations show improved verifier-audited traceability and comparative support in the tested closed-roster setting.

Our contributions are:
\begin{itemize}
    \item a ranking interface whose action couples an ordered candidate list with candidate-linked provenance bundles;
    \item a training objective joining listwise utility, deterministic certificate validity, and evidence-cycle consistency; and
    \item an audit protocol that separates slot traceability, rank-blind reconstruction, held-out-verifier transfer, counterfactual sensitivity, and blinded human judgment.
\end{itemize}

\section{Related Work}
\label{sec:related}

\paragraph{Attribution and faithfulness.}
Attribution metrics ask whether generated claims are supported by cited sources~\citep{ais2023,attributionbench2024,finegrainedcitation2024}; explanation work separately warns that plausible rationales need not be faithful to a model's decision process~\citep{jacovi_faithfully_2020,faithtests2023,causalfaithfulness2025,faithlm2026}.  Candidate prioritization has also been studied in investigative search and triage~\citep{hung_investigative_2016,sutmuller_getting_2020}, where evidence access and accountability are consequential.  ECPO audits an output-level contract: whether source evidence supports recoverable candidate decisions.  The contract evaluates output-level evidence support.

\paragraph{Ranking, planning, and structured generation.}
Learning-to-rank methods optimize comparative order~\citep{burges_ranknet_2010}, while maximum-entropy reward learning~\citep{ziebart_maximum_2008}, plan recognition~\citep{ramirez_plan_2009,ramirez_probabilistic_2010}, and process-conformance/digital-twin models~\citep{brockhoff_process_2021,park_realizing_2021} score partial traces against structured goals.  Constrained decoding improves syntactic validity~\citep{scholak_picard_2021}, and RLHF, preference learning, and group-normalized optimization provide tools for structured LLM outputs~\citep{ouyang_training_2022,schulman_proximal_2017,rafailov_direct_2023,shao_deepseekmath_2024}.  ECPO combines known tools around a different action and evaluation contract: ranking and evidence are generated and audited jointly.

\paragraph{Verifier dependence.}
Verification can improve reasoning and generation, but optimizing directly against a verifier can encourage evaluator-specific behavior~\citep{verifycommit2026,gamingverifiers2026}.  We therefore distinguish the reward verifier from two held-out evaluators, perturb its parameters, reverse the train/evaluate direction, and compare automatic judgments with blinded human annotations.  These tests establish evaluator-form transfer within the shared roster and ontology.

\section{Task and Evaluation Contract}
\label{sec:contract}

For a ranking window $w$, let $\mathcal D_w$ be the exposed documents, $C_w$ a finite candidate roster, $S_w=([s_1,\ldots,s_M],\prec)$ a predefined soft-precedence skeleton, and $\tau_w(c)$ a possibly empty candidate trajectory.  Events carry normalized stages and roles plus zero-indexed half-open provenance spans $(\texttt{doc\_id},[l,r))$.  A candidate is positive when a source-normalized chain supported inside $\mathcal D_w$ is valid and aligned to $S_w$:
\begin{equation}
\mathcal P_w=\{c\in C_w:\exists z\in\mathcal Z^{\rm src}(w,c,S_w),\ \mathrm{valid}(z)=1\},
\qquad \mathrm{rel}_w(c)=\mathbb I[c\in\mathcal P_w].
\end{equation}
The positive set and audit references are supervision/evaluation data and are absent from policy-visible inputs.  The policy sees randomized window-local candidate IDs and emits
\begin{equation}
 y=(A,E),\qquad A=[a_1,\ldots,a_k],\quad E=[E_1,\ldots,E_k],\quad k=\min(10,|C_w|),
\end{equation}
where $E_j=[u_{j1},\ldots,u_{jM}]$ is the ordered, step-aligned evidence bundle paired with $a_j$.  Each $u_{ji}$ cites compatible trigger/argument spans or is the null step $\bot$, serialized exactly as \texttt{matched:false}, \texttt{event\_id:null}, and \texttt{evidence:[]}.  The validator requires strict JSON, in-roster distinct candidates, exactly $k$ positions and $M$ steps per bundle, aligned certificates, valid documents and bounds, and trajectory/stage/role-compatible evidence.

This interface requires only: (i) a finite candidate set supplied or produced separately, (ii) candidate-linked provenance units such as spans, passages, cells, or logs, and (iii) an auditable support contract.  The four-stage event skeleton and frozen extractor are one instantiation.  The evaluation uses a declared provenance and support schema.

\paragraph{Metrics.}
NDCG@10~\citep{jarvelin_cumulated_2002} measures ordinary ranking.  \FeasibleRate{} is the all-window validator pass rate.  \SlotEvid{}@10 is the fraction of submitted slots whose stripped bundle is reassigned to the candidate originally occupying that slot; it measures slot traceability.  Let $z_{w,j}=1$ only when slot $j$ passes the validator and the declared verifier recovers $a_j$.  Verifier-CertNDCG@10 (\vCert{}@10) is
\begin{align}
\mathrm{vCertDCG}@10(w)&=\sum_{j=1}^{k}\frac{(2^{\mathrm{rel}_w(a_j)}-1)z_{w,j}}{\log_2(j+1)},\\
\vCert{}@10(w)&=
\begin{cases}
\mathrm{vCertDCG}@10(w)/\mathrm{IDCG}@10(w),&\mathrm{IDCG}@10(w)>0,\\
0,&\mathrm{IDCG}@10(w)=0.
\end{cases}
\end{align}
It is thus ordinary binary-relevance DCG with a verifier-relative position mask.  The boundedness observation $\mathrm{vCertDCG}\leq\mathrm{DCG}$ follows from the position mask.

For rank-blind evaluation, candidate--bundle pairing and bundle order are additionally removed and bundles are permuted.  A held-out evaluator returns a partial one-to-one assignment and orders recovered candidates by evidence score.  For nonempty submitted order $S=(c_1,\ldots,c_k)$,
\begin{align}
\SetRecover@10 &= |\mathrm{set}(S)\cap\mathrm{set}(\widehat S)|/k,\\
\RankRecover@10 &= k^{-1}\sum_{j=1}^{k}\mathbb I[\widehat S_j=c_j].
\end{align}
Missing, wrong, abstained, and exact-score-tied candidates receive zero positional credit.  Pairwise accuracy scores correct/tied/reversed order over recovered submitted-target pairs as $1/0.5/0$; a window with no recovered pair scores zero.  Kendall's $\tau_b$ retains evidence-score ties after restricting both orders to recovered selected candidates; degenerate windows score zero.  $\AbstainRate{}@10=(k-|\widehat S|)/k$.  All five rank-blind metrics are zero for an empty target set.
Abstention is calibrated rather than maximized: a lower rate is useful only when supported candidates remain accepted, while a higher rate can reflect conservative rejection of unsupported bundles.

\section{Evidence-Coupled Policy Optimization}
\label{sec:method}

\begin{figure}[t]
    \centering
    \includegraphics[width=\linewidth]{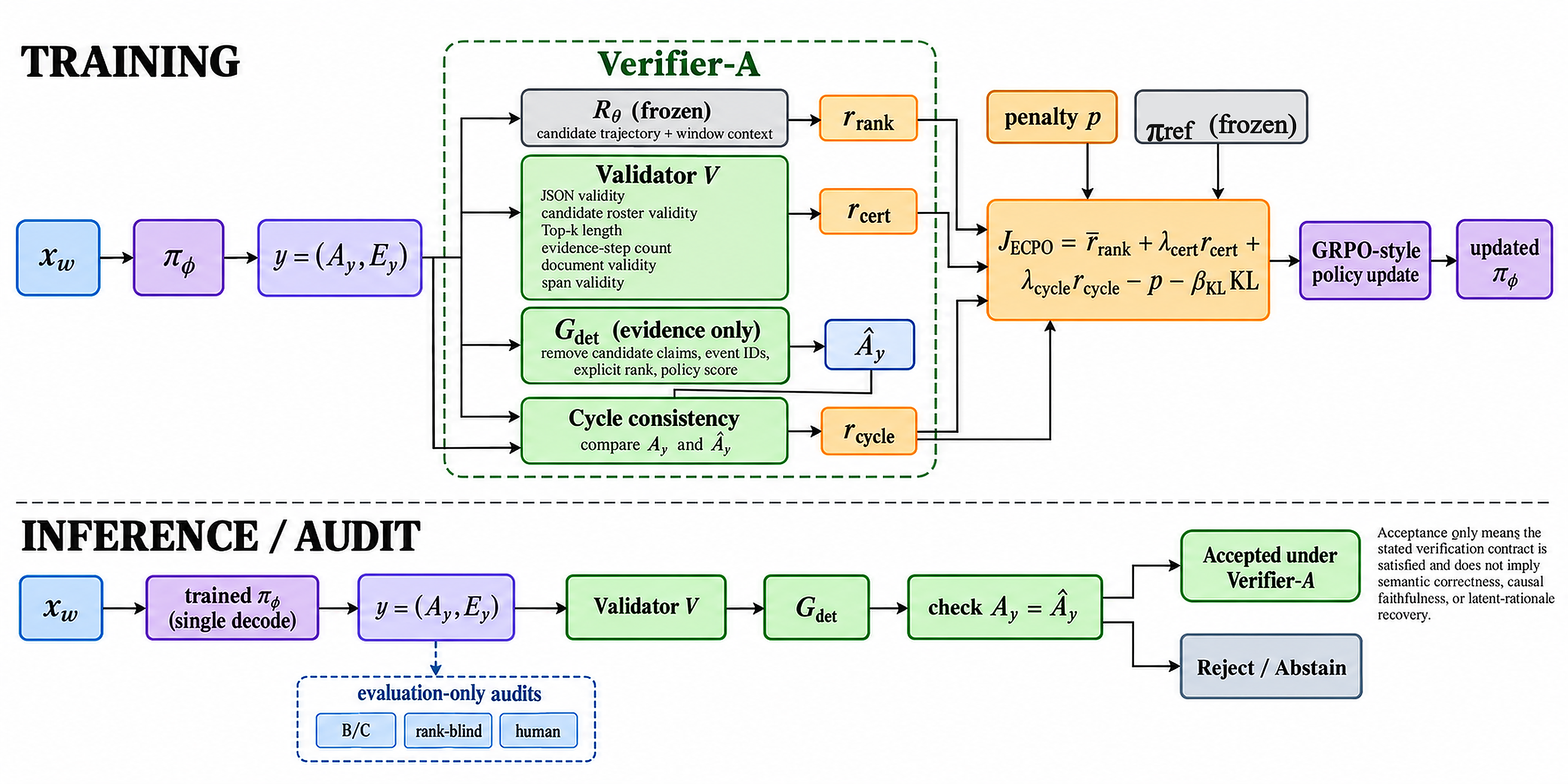}
    \caption{\textbf{ECPO overview.} Ranking and evidence are sampled jointly; the evidence cycle strips answer-bearing fields for candidate recovery.}
    \label{fig:overview}
\end{figure}

Figure~\ref{fig:overview} summarizes training and evaluation.  The policy proposes the joint ranking/certificate object.  A feasibility validator checks the claimed candidate at each position.  The cycle verifier instead removes direct candidate claims, generated event IDs, explicit ranks, and policy scores before scoring evidence against every roster candidate.
Unlike conventional RL post-training, ECPO includes an inference-time audit loop: each generated response is checked against the declared evidence contract before evidence-aware metrics are recorded, with abstention retained as an explicit outcome.

\subsection{Skeleton alignment and trajectory utility}
For trajectory $\tau=[e_1,\ldots,e_T]$, a dynamic program aligns the ordered skeleton to compatible observed events.  Its local score is
\begin{equation}
m(i,t)=\begin{cases}
\mathrm{role\_sat}(s_i,e_t)-\lambda_{\rm arg}\mathrm{arg\_viol}(s_i,e_t),
& \mathrm{stage}(s_i)\in\mathrm{skeleton\_hits}(e_t),\\
-\delta_{\rm type},&\text{otherwise,}
\end{cases}
\end{equation}
Here $\mathrm{skeleton\_hits}(e_t)$ is the deterministic set of compatible stages assigned from the event type and trigger.  If $\mathcal R_i$ is the required-role set and $\mathcal A_t$ the normalized arguments of $e_t$, then
\begin{align}
\mathrm{role\_sat}(s_i,e_t)&=
\begin{cases}1,&|\mathcal R_i|=0,\\
|\mathcal R_i|^{-1}\sum_{r\in\mathcal R_i}\mathbb I[r\text{ occurs in }\mathcal A_t],&\text{otherwise,}
\end{cases}\\
\mathrm{arg\_viol}(s_i,e_t)&=\sum_{r\in\mathcal R_i}
\mathbb I[r\text{ violates a declared role/entity constraint in }e_t].
\end{align}
Main runs check candidate identity for key roles and available role-specific incompatibilities, but impose no extra entity-type constraint.  Precedence is a soft post-backtracking feature, not a local-match term.  Match, missing-step, and skipped-event transitions yield coverage, role, precedence, temporal, and optional graph features $\phi(\tau;S_w)$, with $R_\theta(\tau;S_w)=\theta^\top\phi(\tau;S_w)$.  We learn $\theta$ with a window-wise maximum-entropy likelihood plus Bradley--Terry preferences over train-split positive, same-window, and perturbed trajectories.  Appendix~\ref{app:align} gives the recurrence, penalties, tie rules, and reward-model objective.

\subsection{Verifier-A}
\label{sec:verifier-a}
For cited and event spans in the same document, Verifier-A uses
\begin{equation}
 \mathrm{IoU}([l_s,r_s),[l_e,r_e))=
 \frac{\max(0,\min(r_s,r_e)-\max(l_s,l_e))}
 {\max(r_s,r_e)-\min(l_s,l_e)}.
\end{equation}
A match requires strict $\mathrm{IoU}>0.50$; equality fails.  Compatible triggers are tested first.  Arguments are considered only if no trigger matches and additionally require normalized-role agreement.

For stripped bundle $b$ and candidate $c$, the declared support score is
\begin{equation}
 A(b,c)=0.30\,\mathrm{StepCov}+0.20\,\mathrm{RoleSat}
 +0.25\,\mathrm{Trace}+0.10\,\mathrm{Prec}-0.15\,\mathrm{BadSpan}.
 \label{eq:verifier-a}
\end{equation}
StepCov is the fraction of all $M$ steps with valid, traceable, stage-compatible support (zero if $M=0$).  RoleSat is the fraction of required $(\text{step},\text{role})$ instances supported, and equals one when none are required.  Trace is the fraction of citations traceable to $c$ and is zero for an empty bundle.  Prec is the fraction of evaluable supported precedence edges that are consistent and is zero if none are evaluable.  BadSpan is the fraction of citations with any invalid, duplicate, off-window, wrong-role, type-incompatible, non-traceable, or multi-candidate-ambiguous defect; each citation counts at most once and an empty bundle scores zero.

The weights encode priorities fixed before policy training and output inspection: coverage and traceability receive $0.55$ jointly, role support $0.20$, precedence $0.10$ because order metadata can be unreliable, and defective spans $-0.15$.  At the $0.55$ support threshold, no single positive component can pass; the $0.10$ margin rejects near-ties.  These choices used no development/test relevance labels.  Appendix~\ref{app:sensitivity} evaluates 47 alternatives on fixed outputs.

Assignment requires $A(b,c_{\rm best})\geq0.55$ and $A(b,c_{\rm best})-A(b,c_{\rm second})\geq0.10$; equality passes and failure of either test abstains.  A maximum-weight one-to-one assignment prevents reuse.  Submitted slot-wise evaluation resolves ties by bundle order and then window-local candidate order.  Rank-blind evaluation instead abstains on exact assignment ties and never uses bundle order; equal order scores remain ties.  This distinction is essential to permutation invariance.

\subsection{Evidence-coupled reward}
For a parsed output $y=(A,E)$ with $L=|A|$, define
\begin{align}
r_{\rm rank}(y,w)&=\sum_{j=1}^{L}\gamma^{j-1}R_\theta(\tau_w(a_j);S_w),\\
\overline r_{\rm rank}(y,w)&=
\frac{r_{\rm rank}(y,w)-\mu_{\rm rank}(w)}
{\sigma_{\rm rank}(w)+\epsilon_{\rm rank}},\\
r_{\rm cert}(y,w)&=\frac1k\sum_{j=1}^{k}
\mathbb I[V(E_j,a_j,w)=1]\,\mathrm{StepCov}(E_j,S_w,a_j),\\
r_{\rm cycle}(y,w)&=\frac1k\sum_{j=1}^{k}\mathbb I[\hat a_j=a_j].
\end{align}
The first term is normalized within the window's sampled group; $V$ is the feasibility validator, and $\hat a_j$ is Verifier-A's one-to-one assignment read at submitted slot $j$ after answer-bearing fields are stripped.  Indicators for missing positions $j>L$ are defined as zero.  Thus certificate validity and evidence-only candidate recovery are separate signals.
Here $r_{\rm rank}$ is supplied by the learned trajectory reward, $r_{\rm cert}$ measures validator-backed certificate support, and $r_{\rm cycle}$ measures recovery after evidence reconstruction.  The three terms therefore have distinct sources and purposes: preference optimization, evidence validity, and decision--evidence consistency.  The scalar reward is
\begin{equation}
r_{\ECPO}=\overline r_{\rm rank}+\lambda_{\rm cert}r_{\rm cert}
+\lambda_{\rm cycle}r_{\rm cycle}-\xi_{\rm inv}\mathbb I_{\rm invalid}
-\xi_{\rm miss}(k-|A|)_+-\beta_{\rm KL}\widehat{\mathrm{KL}}_{\rm tok}.
\end{equation}
Unparseable outputs and outputs without a valid candidate prefix receive zero component utilities plus invalid/missing penalties; shorter parsed prefixes are scored only at present positions and penalized for the deficit.  The main runs use discount $0.90$, $\lambda_{\rm cert}=0.5$, $\lambda_{\rm cycle}=1.0$, both penalties $2.0$, group size 8, clip range $0.2$, and initial $\beta_{\rm KL}=0.05$.  A Qwen3-8B SFT policy is updated with GRPO-style group-normalized advantages~\citep{shao_deepseekmath_2024}; $R_\theta$ and Verifier-A are fixed during each policy phase.  Appendix~\ref{app:optimization} gives the token-level KL estimator and clipped objective.  This is an evidence-coupled formulation built from established optimization components.

\section{Experimental Protocol}
\label{sec:protocol}

\paragraph{Primary datasets.}
MAVEN-ERE and RAMS supply source-normalized labels and audit references, while every ranking method consumes separate frozen-extractor trajectories.  Ranking windows follow dataset-native document or document-local grouping, not calendar slicing.  When no external roster is available, a label-free resolver fixes $C_w$ from an identifier-only projection of source-normalized entities in agent-like or target-like roles; contextual entities enter only if they also occupy a candidate role.  This projection retains anonymized identity and window membership but removes stage/chain validity, scores, positive labels, and reference spans before frozen predictions attach trajectories.  Candidate-participating predicted events are ordered by timestamp when available and otherwise by dataset/document order.  MAVEN-ERE compact contexts use one trigger sentence plus at most one neighbor on each side (no stride); RAMS extractor/SFT contexts use the first 120 characters, whereas ranking prompts retain full provenance pointers.  Document-level splits prevent a document from crossing train/dev/test.  The evaluated construction contains 576/71/71 MAVEN-ERE and 716/90/90 RAMS windows; test windows average 10.1 and 9.7 candidates.  Policy-visible IDs are randomized per window.  The primary closed roster isolates ranking/evidence behavior; predicted and hybrid rosters are availability diagnostics.

\paragraph{Models and matched controls.}
Primary policies share the Qwen3-8B backbone~\citep{yang_qwen3_2025}, QLoRA adaptation~\citep{qlora2023}, SFT initialization, prompt/schema, candidate pools, and single-decode budget.  Plain GRPO uses train-label ranking utility plus schema, candidate-ID, and evidence-format validity; it uses neither $R_\theta$ nor a DP-alignment bonus.  Besides SFT and GRPO, we emphasize two matched controls.  RM-only+Det ranks with $R_\theta$ and searches validator-valid evidence after fixing the ranking.  GRPO+Cert is separately trained with the plain-GRPO reward plus the same non-cycle $r_{\rm cert}$ and compute budget, but omits $r_{\rm cycle}$.  The $-r_{\rm cycle}$ row is the exact within-ECPO mechanism ablation.  Main values are test means; ECPO uses three seeds, and paired 95\% intervals resample source windows, the inference and bootstrap cluster.

\paragraph{External validity tests.}
Verifier-B is deterministic and has no learned component: it replaces IoU with citation coverage, retains StepCov/Trace/BadSpan, removes RoleSat/Prec, and sets parameters from disjoint train-split synthetic corruptions without dev/test labels, policy outputs, rewards, or audit references.  Verifier-C is a frozen policy-held-out raw-text evaluator that re-extracts entity, stage, and role from each citation plus $\pm200$ characters, maps entities to roster candidates through window-local aliases/coreference, assigns one-to-one with abstention, and orders by evidence-derived support.  C's training/calibration uses only train-split raw-text/span annotations---never generated certificates, rankings, policy/reward outputs, or test documents.  Unlike \ReExtractFaithfulness{}, which only audits per-span stage/role compatibility, C reconstructs candidate identity and order.  Neither B nor C affects reward, checkpoint selection, decoding, or test thresholds for primary ECPO-A.  A separate reverse-transfer ECPO-B policy uses B in its reward and is evaluated with A/C.  We additionally test Llama-3.1-8B-Instruct~\citep{meta_llama31_2024}, Qwen3-14B~\citep{yang_qwen3_2025}, and HotpotQA~\citep{hotpotqa2018}.  HotpotQA uses an answer roster from shared retrieved context; its retained protocol is 740 dev/740 test examples with Qwen3-8B-Instruct, seeds $(42,43,44)$, greedy decoding with maximum output 768, and 2000 paired bootstrap replicates.

\section{Results and Validity}
\label{sec:results}

\begin{table}[htbp]
\caption{\textbf{Primary utility and matched controls at $K=10$.} Panel values are MAVEN-ERE (M) or RAMS (R). \vCert$_A$ and \SlotEvid{} use Verifier-A; \RankRecover$_B$ is rank-blind held-out-Verifier-B evaluation. Bold and underlined entries mark the best and second-best values within each dataset/metric.}
\label{tab:main}
\centering
\small
\begin{tabular}{@{}llccccc@{}}
\toprule
Data & Method & NDCG & \SlotEvid{} & \vCert$_A$ & \shortstack{Rank\\Recover$_B$} & Feas.\\
\midrule
M & SFT & 0.61 & 0.36 & 0.31 & 0.27 & 0.77\\
M & GRPO & 0.65 & 0.44 & 0.39 & 0.34 & 0.82\\
M & GRPO+Cert & 0.67 & 0.50 & 0.45 & 0.40 & 0.90\\
M & RM-only+Det & \best{0.72} & \second{0.53} & \second{0.49} & 0.41 & \best{0.98}\\
M & ECPO $-r_{\rm cycle}$ & 0.68 & 0.49 & 0.43 & \second{0.46} & 0.93\\
M & \best{ECPO} & \second{0.70} & \best{0.65} & \best{0.58} & \best{0.55} & \second{0.94}\\
\midrule
R & SFT & 0.53 & 0.25 & 0.21 & 0.19 & 0.71\\
R & GRPO & 0.57 & 0.32 & 0.27 & 0.24 & 0.78\\
R & GRPO+Cert & 0.59 & 0.38 & 0.33 & 0.29 & 0.87\\
R & RM-only+Det & \best{0.63} & \second{0.40} & \second{0.36} & 0.30 & \best{0.97}\\
R & ECPO $-r_{\rm cycle}$ & 0.59 & 0.34 & 0.29 & \second{0.35} & 0.91\\
R & \best{ECPO} & \second{0.61} & \best{0.49} & \best{0.40} & \best{0.42} & \second{0.92}\\
\bottomrule
\end{tabular}
\end{table}

\paragraph{Utility--audit trade-off.}
Table~\ref{tab:main} makes the trade-off explicit.  ECPO-A reaches \vCert{}@10 of $0.58/0.40$ and \SlotEvid{}@10 of $0.65/0.49$, but RM-only+Det has higher ordinary NDCG ($0.72/0.63$ versus $0.70/0.61$).  Removing $r_{\rm cycle}$ changes \vCert{} from $0.58$ to $0.43$ and $0.40$ to $0.29$, while NDCG changes only from $0.70$ to $0.68$ and $0.61$ to $0.59$.  This isolates the effect of evidence coupling.

\begin{table}[H]
\caption{\textbf{Construct-validity audits} on MAVEN-ERE/RAMS. Panel A reports \vCert{} under evaluators A/B/C; Panel B removes positions and candidate--bundle pairing. Slash-separated entries are M/R; scalar entries are pooled retained diagnostics.}
\label{tab:construct}
\centering
\small
\begin{tabular}{@{}lccc@{}}
\toprule
\multicolumn{4}{@{}l}{Panel A: cross-verifier \vCert{}@10 (M/R)}\\
Training run & Eval. A & Eval. B & Eval. C\\
\midrule
GRPO+Cert & 0.45/0.33 & 0.41/0.28 & 0.40/0.27\\
\best{ECPO-A} & \best{0.58/0.40} & \best{0.55/0.38} & \best{0.53/0.36}\\
\second{ECPO-B} & \second{0.56/0.39} & \second{0.54/0.37} & \second{0.52/0.36}\\
\bottomrule
\end{tabular}

\vspace{1pt}
\begin{tabular}{@{}lccccc@{}}
\toprule
\multicolumn{6}{@{}l}{Panel B: rank-blind recovery (M/R)}\\
Method/eval. & SetRec. & RankRec. & PairAcc. & $\tau_b$ & AbstainRate\\
\midrule
GRPO/B & 0.50/0.37 & 0.34/0.24 & 0.62/0.56 & 0.31/0.21 & \best{0.08}\\
RM-only+Det/B & 0.58/0.43 & 0.41/0.30 & 0.66/0.60 & 0.39/0.28 & 0.22/0.32\\
\best{ECPO-A/B} & \best{0.70/0.55} & \best{0.55/0.42} & \best{0.75/0.69} & \best{0.53/0.41} & \best{0.16/0.24}\\
\second{ECPO-A/C} & \second{0.67/0.52} & \second{0.52/0.39} & \second{0.73/0.67} & \second{0.49/0.38} & \second{0.12}\\
\bottomrule
\end{tabular}
\end{table}

\paragraph{Held-out and rank-blind evaluation.}
Table~\ref{tab:construct} shows that primary ECPO-A retains \vCert{}@10 of $0.55/0.38$ under B and $0.53/0.36$ under C.  Its advantage over RM-only+Det is $+0.07[0.03,0.11]/+0.04[0.01,0.07]$ under B and $+0.06[0.02,0.10]/+0.03[0.01,0.06]$ under C.  We report the retained paired gains for these evaluations.  Reverse-transfer ECPO-B obtains $0.56/0.39$ under A and $0.52/0.36$ under C.  In rank-blind reconstruction, ECPO-A improves \RankRecover{} over RM-only+Det by $+0.14[0.08,0.20]$ and $+0.12[0.06,0.18]$ while abstaining less.  The rank-blind audit measures output recovery.

\paragraph{Counterfactual evidence.}
Same-candidate random spans retain target recovery of $0.52/0.40$, confirming that candidate identity can sometimes be recovered from arbitrary candidate-linked evidence.  Yet \RankRecover{} falls from $0.55/0.42$ to $0.25/0.18$, and Verifier-B \vCert{} falls from $0.55/0.38$ to $0.30/0.21$.  Replacing one bundle with a length- and stage-matched valid bundle for an unselected same-window candidate reduces original-target recovery to $0.14/0.11$.  Appendix~\ref{app:counterfactual} reports all controls.  These results motivate our separation of traceability from comparative support.

\begin{table}[H]
\caption{\textbf{External validation} (MAVEN-ERE/RAMS unless stated).  Panel A is the blinded human audit.  Panel B reports matched within-setting \vCert{}@10 under each setting's policy-held-out evaluator.}
\label{tab:external}
\centering
\small
\begin{tabular}{@{}lccc@{}}
\toprule
\multicolumn{4}{@{}l}{Panel A: human audit (M/R)}\\
Method & Comparative support & Set support & Rank support\\
\midrule
GRPO & 0.55/0.47 & 0.51/0.39 & 0.36/0.27\\
RM-only+Det & \second{0.64/0.59} & \second{0.60/0.46} & \second{0.43/0.32}\\
\best{ECPO} & \best{0.76/0.69} & \best{0.72/0.56} & \best{0.57/0.44}\\
\bottomrule
\end{tabular}

\vspace{1pt}
\begin{tabular}{@{}lccc@{}}
\toprule
\multicolumn{4}{@{}l}{Panel B: backbone/schema transfer}\\
Setting and baseline & Baseline & ECPO & Gain\\
\midrule
Llama-3.1-8B-Instruct, M / GRPO & 0.37 & \best{0.49} & +0.12\\
Llama-3.1-8B-Instruct, M / RM-only+Det & 0.42 & \best{0.49} & +0.07\\
Llama-3.1-8B-Instruct, R / GRPO & 0.26 & \best{0.35} & +0.09\\
Llama-3.1-8B-Instruct, R / RM-only+Det & 0.31 & \best{0.35} & +0.04\\
Qwen3-14B, M / GRPO & 0.41 & \best{0.53} & +0.12\\
Qwen3-14B, R / GRPO & 0.28 & \best{0.38} & +0.10\\
HotpotQA / GRPO & 0.31 & \best{0.39} & +0.08\\
HotpotQA / RM-only+Det & 0.35 & \best{0.39} & +0.04\\
\bottomrule
\end{tabular}
\end{table}

\paragraph{Human support and transfer.}
The blinded audit covers 120 stratified test windows and 1,080 bundles per method (540 per dataset), totaling 3,240 base method--bundle units.  Annotators see neither method, scores, labels, stable IDs, nor ranks; ambiguity counts as failure.  ECPO's human comparative support is $0.76/0.69$, a gain over RM-only+Det of $+0.12[0.07,0.17]/+0.10[0.05,0.15]$.  Comparative-support agreement is Krippendorff's $\alpha=0.71$, and a stratified 20\% subset (648/3,240 units) is independently repeat-annotated.  Table~\ref{tab:external} also shows persistence across Llama-3.1-8B-Instruct, Qwen3-14B, and the HotpotQA two-hop provenance contract.  On HotpotQA, the \vCert{} gain is $+0.08[0.03,0.12]$ over GRPO and $+0.04[0.01,0.07]$ over RM-only+Det.  These tests show gap persistence in the evaluated backbones and provenance-preserving schemas.

\paragraph{Sensitivity, extractor ceilings, and cost.}
Across 47 evaluation-only Verifier-A configurations, \vCert{} ranges over $0.55$--$0.60$ on MAVEN-ERE and $0.38$--$0.42$ on RAMS; method ordering is preserved in 45/47 and 43/47 settings.  Fixed-roster predicted-evidence oracles reach $0.72/0.54$, source-normalized oracles $0.86/0.76$, a higher-recall extractor $0.64/0.48$, and 20\% predicted-event deletion $0.50/0.33$.  Missing upstream support cannot receive verifier credit.  Beyond the shared extractor, the reward model costs 8 GPU-hours.  RM-only+Align takes 0.18 s/window; ECPO adds 76 GPU-hours of policy optimization per corpus-seed and takes 3.1 s/window for one decode.  NDCG and latency remain explicit trade-offs.

\section{Limitations}
\label{sec:limitations}
All primary experiments use a closed finite roster, a fixed extractor, candidate-linked provenance, predefined skeleton/role semantics, and an automatic support contract. Verifier-B and Verifier-C change evaluator form while retaining the roster and broad ontology. The two primary datasets are event-centric, and HotpotQA adds one provenance schema. Automatic scores are verifier-relative, the human audit is finite, upstream representation bounds available evidence, and ECPO has higher cost than deterministic ranking.

\section{Conclusion}
We presented an evidence-coupled ranking interface, training objective, and audit protocol.  In the tested closed-roster provenance settings, ECPO improves verifier-audited traceability and rank-blind and human-assessed comparative support relative to matched controls, including under held-out evaluators.  The stronger deterministic ranker's ordinary NDCG and cost make the deployment trade-off explicit: ECPO is appropriate when ranking and evidence must be generated as one auditable object.

\subsection*{AI use statement}
Generative AI tools were used to assist with LaTeX template migration, manuscript organization, language editing, reference formatting, and literature-search assistance.  The scientific content, including the method, experimental design, numerical results, and their interpretation, was supplied and verified by the authors; generative AI was not used to generate data or implement the method.  The authors reviewed all AI-assisted content and take responsibility for the final work.

\subsection*{Ethics statement}
The completed blinded audit sampled 120 stratified test windows, covering 1,080 bundles per method and 3,240 method--bundle units overall.  A stratified 20\% subset (648/3,240) was independently repeat-annotated, with comparative-support Krippendorff's $\alpha=0.71$.  Annotators were blinded to method identity, model/evaluator scores, labels, stable IDs, and submitted ranks; ambiguity counted as failure.  They received written instructions and risk/confidentiality information, were paid a fixed hourly rate above the local minimum wage, and received no performance bonus.  An institutional minimal-risk determination found that full IRB review was not required.  ECPO is intended for decision support with human review, access controls, logging, bias/drift audits, and appeal or correction channels in sensitive deployments.

\subsection*{Reproducibility statement}
Sections~\ref{sec:contract}--\ref{sec:protocol} define the task, verifier, objective, baselines, and split protocol. Appendices~\ref{app:data-construction}, \ref{app:align}, \ref{app:metric-semantics}, and \ref{app:artifact-checklist} give data construction, exact edge cases, statistical protocols, and the artifact completion ledger.

\bibliographystyle{iclr2027_conference}
\bibliography{iclr2027_conference_restored}

\appendix
\raggedbottom
\section{Detailed Task, Data, and Construction Protocol}
\label{app:data-construction}

This appendix restores the task and data details needed to reconstruct the evaluated closed-roster interface. The current ICLR specification is authoritative: a policy-visible record contains no relevance label, source-normalized chain, audit reference, stable cross-window candidate identifier, or verifier answer. The appendix records the policy-visible fields and evaluation contract.

\paragraph{Source separation.} For every window, three views remain distinct.  The \emph{source-normalized view} supplies train-only supervision and held-out relevance/audit references.  The \emph{model-input view} is produced by a frozen extractor and supplies the trajectories visible to every ranking method.  The \emph{policy-output view} contains only the generated ordered roster IDs and evidence bundles.  No source-normalized span, chain-validity flag, or candidate label is copied into a policy prompt.  The evaluator may join the views by hidden window and candidate keys only after the output has been fixed.

\paragraph{Compatibility rule for restored statistics.} The descriptive statistics below are inherited only where the old and current manuscripts declare the same datasets, document-level split discipline, window construction, candidate resolver, and frozen-extractor interface. They are independent of the changed Verifier-A score.

\subsection{Problem framing details}
\label{sec:app_moved_intro_details}

This subsection expands the brief main-text discussion of the intent, skeleton, and window-graph abstractions.

\paragraph{Ranking windows and context snippets.}
A ranking window is the evidence scope for one Top-$K$ candidate-ranking decision.
It is not necessarily a calendar-time interval.
A context snippet is a local text segment used for upstream extraction or compact prompting.
In MAVEN-ERE, source-record snippets used for supervised extractor examples and compact evidence display are trigger-centered sentence neighborhoods.
In RAMS, compact source snippets are document prefixes.
Held-out extractor inference does not use gold trigger locations: the frozen extractor is applied to split documents or document-local text units, and predicted event spans are linked back to their surrounding local context for prompt compaction and auditing.
The number of snippets and documents contributing to each ranking window is an empirical property reported in Appendix~\ref{sec:app_moved_data_details} and Appendix Table~\ref{tab:ranking_window_stats}.

\paragraph{Intent and plan skeleton.}
In this paper, an \emph{intent} denotes a hypothesized multi-step activity type that we instantiate with an explicit plan skeleton.
In our benchmark instantiations, we use a four-stage inventory
\texttt{PREP}$\rightarrow$\texttt{PROBE}$\rightarrow$\texttt{EXECUTE}$\rightarrow$\texttt{OUTCOME}
as the canonical soft order.
Individual \texttt{intent\_id} skeleton records instantiate ordered steps, role requirements, and optional soft-precedence edges over this inventory.
We operationalize an intent with an explicit \emph{plan skeleton} $S$, which is an \emph{intent-level} abstraction shared by all windows labeled with the same \texttt{intent\_id}.
Importantly, $S$ is \emph{not} constructed by directly taking any single candidate's observed event sequence as the skeleton; instead, it specifies high-level step \emph{types} and key role constraints, and individual candidates are only partially observed realizations that may miss steps, contain spurious events, or have uncertain timestamp / document-order indices.
Ordering irregularities are captured by soft precedence-violation features after monotone alignment, not by arbitrary event transpositions in the DP recurrence.
In the main benchmark setting, each \texttt{intent\_id} is associated with a single predefined soft-precedence plan skeleton.
We allow missing steps and skipped events during alignment to handle partial observability.
To examine the effect of this assumption, Appendix~\ref{sec:app_hard_windows} reports diagnostic skeleton-perturbation analyses.
When reported, Appendix~\ref{sec:app_hard_windows} treats skeleton banks only as an auxiliary diagnostic; they are not part of the main method or main-result comparison.
Automatically discovering skeletons from raw text remains outside the scope of this paper.
Table~\ref{tab:skeleton_example} shows an illustrative skeleton record; it is schematic and does not expose any held-out labels or audit references.

\begin{table}[htbp]
\centering
\caption{Illustrative predefined skeleton record. Stage labels are abstract alignment labels; \texttt{OUTCOME} denotes the terminal or outcome stage in the declared JSON schema.}
\label{tab:skeleton_example}
\small
\setlength{\tabcolsep}{4pt}
\begin{tabularx}{\linewidth}{l l Y}
\hline
Step & Stage & Required roles / soft-precedence description \\
\hline
$s_1$ & \texttt{PREP} & optional \texttt{Agent}; precedes later evidence-bearing steps when order is available \\
$s_2$ & \texttt{PROBE} & \texttt{Agent} and optional \texttt{Target}; soft-precedes execution evidence \\
$s_3$ & \texttt{EXECUTE} & \texttt{Agent} and \texttt{Target}; central action step for alignment \\
$s_4$ & \texttt{OUTCOME} & optional \texttt{Target}/\texttt{Context}; terminal or outcome evidence \\
\hline
\end{tabularx}
\end{table}

Additional details on candidate identities, shared documents, and evidence-span serialization are provided in Appendix~\ref{sec:app_moved_candidate_identity}.

This paper focuses on the ranking-and-audit interface itself: candidate prioritization is evaluated only through window-local evidence, randomized candidate identifiers, and auditable span certificates.

\subsection{Candidate identities, document sharing, and evidence serialization}
\label{sec:app_moved_candidate_identity}

In this paper, a \emph{candidate} is an anonymized entity cluster placed into a window-local candidate pool by a label-free resolver over role-bearing event arguments.
When no external roster is available, the main benchmark first fixes an eligible closed roster from an identifier-only source projection of role-bearing entities.
The resolver then links frozen-extractor predicted mentions only to this eligible roster; predicted-only entities absent from the closed roster are excluded from the main benchmark and are analyzed separately in the predicted-only and hybrid-roster diagnostics.
The candidate pool $C_w$ is local to a ranking window. For all policy prompts and generated outputs, candidate identifiers are randomized and window-local (for example, \texttt{cand\_001}); source-normalized entity IDs are retained only in hidden evaluator metadata.
Ranking, validation, and metric computation therefore compare candidates within the current $C_w$ and do not allow cross-window ID memorization.
Identifier controls, pooled over both datasets, are reported in Appendix~\ref{sec:app_id_controls}.
Although motivating applications often involve persons of interest, MAVEN-ERE and RAMS contain heterogeneous event participants; therefore, the declared benchmark uses candidate entities after deterministic role normalization.
Candidate pools are constructed from entities that participate in agent-like or target-like roles, while purely contextual entities are not used as candidates unless they also appear in a candidate role.

A document may mention multiple candidates; therefore, the same document $d\in\mathcal{D}_w$ can contribute evidence spans to multiple candidates.
Likewise, a single predicted event record may appear in multiple candidate-centric trajectories if multiple candidates participate in its arguments.
Each trajectory remains auditable via its own \texttt{doc\_id:span} references.

We represent an evidence span conceptually as $\sigma=(\texttt{doc\_id},[l,r))$, where $[l,r)$ is a 0-indexed half-open character-offset interval.
In JSON, the bounds are serialized either as the two-integer field \texttt{span:[l,r]} or as the string \texttt{"l-r"}.
In both cases, the stored pair $(l,r)$ is interpreted by the validator as the same half-open interval $[l,r)$; the array notation is a serialization of endpoints rather than a closed mathematical interval.

\subsection{Detailed problem notation and event representation}
\label{sec:app_problem_details}

This subsection collects notation and serialization details moved out of the main text.
For a ranking window $w$, $C_w=\{c_{w,1},\ldots,c_{w,N_w}\}$ denotes the window-local candidate pool, and $\tau_w(c_{w,i})\equiv\tau_{w,i}$ denotes the candidate trajectory within $w$.
If no predicted event is available for a roster candidate, we retain an empty trajectory record rather than dropping the candidate.
Events are sorted by absolute time when available and otherwise by dataset-native or document-local order.

\begin{table}[htbp]
\centering
\caption{Core notation.}
\label{tab:notation}
\small
\setlength{\tabcolsep}{4pt}
\begin{tabularx}{\linewidth}{lY}
\hline
Symbol & Meaning \\
\hline
$w$ & ranking window / evaluation instance \\
$\texttt{intent\_id}(w)$ & intent identifier for window $w$ \\
$S_w$ & predefined skeleton associated with $\texttt{intent\_id}(w)$ \\
$C_w$ & candidate pool in window $w$ \\
$N_w$ & number of candidates, $|C_w|$ \\
$\tau_w(c_{w,i})\equiv \tau_{w,i}$ & trajectory of candidate $c_{w,i}$ within window $w$ \\
$K_w$ & evaluated cutoff, $\min(K,N_w)$ \\
$\phi(\tau;S)$ & skeleton, role, precedence, and temporal feature vector \\
$R_\theta(\tau;S_w)$ & learned skeleton-conditioned trajectory-level reward \\
$y$ & generated JSON output containing Top-$K_w$ IDs and certificates \\
\hline
\end{tabularx}
\end{table}

Each event is represented as
\begin{equation}
e =
\big(
\mathrm{etype}^{\mathrm{raw}},
\mathcal{H}^{\mathrm{skel}},
\mathrm{etype}^{\mathrm{primary}},
\sigma^{\mathrm{trig}},
\mathcal{A},
t
\big),
\end{equation}
where $\mathrm{etype}^{\mathrm{raw}}$ is the extractor's fine-grained event type, $\mathcal{H}^{\mathrm{skel}}=\texttt{skeleton\_hits}(e)$ is the set of compatible skeleton stages, and $\mathrm{etype}^{\mathrm{primary}}$ is a single primary stage stored only for serialization and analysis.
Alignment and reward computation use $\texttt{skeleton\_hits}(e)$ rather than the primary stage label.
The argument set $\mathcal{A}$ contains tuples $(\texttt{role},\texttt{entity\_id},\sigma^{\mathrm{arg}})$.
The temporal field $t$ is an absolute timestamp when available, and otherwise a deterministic relative or document-local order key.
Serialized trajectory records may expose both an optional \texttt{time} field and a required \texttt{order\_index}; sorting prioritizes reliable absolute time and falls back to \texttt{order\_index}.
All triggers and arguments preserve auditable provenance via their spans $\sigma$.

For target cutoff $K$, the evaluated list length is $K_w=\min(K,N_w)$.
When a candidate pool contains fewer than $K$ candidates, the expected output contains all candidates in that pool.
Each returned candidate has one certificate object containing one serialized step object for every skeleton step in $S_w$; matched steps cite supporting \texttt{doc\_id:span} evidence and unmatched steps are recorded explicitly with empty evidence.

\subsection{Dataset construction and preprocessing details}
\label{sec:app_moved_data_details}

\paragraph{Relation to the condensed main-text setup.}
For space, the main text compresses task instantiation, datasets, extractor pretraining, reward-learning corpus construction, models/baselines, and metrics into a single experimental-setup subsection. This appendix, together with Appendix~\ref{sec:app_irl_corpus} and Appendix~\ref{sec:app_extractor}, provides the corresponding full details.

\paragraph{Upstream event supervision (extractor only).}
Event extraction is an upstream preprocessing step.
We pretrain the extractor with a staged curriculum:
\texttt{MAVEN} $\rightarrow$ \texttt{MAVEN-Arg} $\rightarrow$ \texttt{MAVEN-ERE} $\rightarrow$ \texttt{RAMS}
to improve argument-role consistency and reduce trajectory noise; details are in Appendix~\ref{sec:app_extractor}.
This curriculum affects only the extractor; the \emph{ranking benchmark instantiations} in this paper are built on \textbf{MAVEN-ERE} and \textbf{RAMS}.

\paragraph{Two evaluation instantiations.}
We instantiate the ranking benchmark in two variants derived from the source corpora \textbf{MAVEN-ERE} and \textbf{RAMS}.
Source-normalized records from these corpora define supervision, positive labels, and audit references, while model-input trajectories are constructed separately from frozen-extractor predictions.
For each ranking window $w$, we aggregate predicted event records by candidate identity, normalize timestamps when available, and otherwise preserve dataset-native or document-local order.
This yields a possibly empty predicted trajectory $\tau_{w,i}$ with auditable provenance \texttt{doc\_id:span} for triggers and arguments.

\paragraph{Windows and trajectory instances.}
Each evaluation instance is a \emph{ranking window} $w$ that defines the evidence scope $\mathcal{D}_w$, the window-local candidate pool $C_w$, source-normalized positive labels, audit references, and model-input trajectories for one Top-$K$ ranking prompt.
A trajectory is \emph{window-scoped}: it is defined for a specific $(w,c)$ pair, i.e., $\tau_{w,c}$ aggregates candidate-participating events that are observable within $\mathcal{D}_w$ and orders them by time or by dataset-native / document-local order when absolute time is unavailable.

We distinguish two event-record views during construction.
Source-normalized records are deterministic conversions of the dataset annotations.
They are used to derive positive candidate labels and reference evidence for audit metrics.
When no external candidate roster is available, we construct the benchmark candidate roster from an identifier-only projection of source-normalized role-bearing entities before relevance labels are attached.
This roster-construction step keeps only anonymized entity identifiers and document/window membership needed to define the candidate universe.
It discards event-stage labels, skeleton-chain validity, positive labels, reward scores, and reference evidence chains.
Thus, source-normalized records may define which anonymized entities are eligible candidates, but they do not provide model-input trajectories or ranking evidence.
Predicted records are produced by the frozen event extractor and are the only event records consumed by the reward model, ranking policy, RM-only ranker, and structured baselines.
The candidate pool $C_w$ is fixed before positive labels are attached, and positive candidate IDs are stored only in \texttt{window\_label.jsonl}; they are not included in model-input records.

The same underlying source entity may appear in multiple ranking windows when a cross-window identifier is available.
However, ranking decisions are made over window-local candidate IDs.
Thus, the number of window-scoped trajectory instances can exceed the number of unique source entities, and overlapping windows may reuse portions of the same document evidence.
Appendix Algorithm~\ref{alg:window_construction} summarizes the construction procedure.

\paragraph{Instantiation-specific context snippets.}
\textbf{MAVEN-ERE.} We construct one fixed \emph{trigger-centered context snippet} per event record used for extraction or compact prompting (no sliding stride): we locate the sentence containing the trigger and take at most three sentences in total (the trigger sentence plus up to one sentence before and one after; \texttt{context\_radius}=1).
\textbf{RAMS.} Upstream event records use dataset-native context construction.
For extractor/SFT text fields, we flatten sentences into a document string and keep its first 120 characters as the local context snippet.
Ranking-policy prompts additionally include the \texttt{intent\_id}, skeleton, candidate IDs, and compact trajectory summaries or evidence pointers.

The full construction (including fallbacks when sentence boundaries are missing) and leakage-preventing splits are described in Appendix~\ref{sec:app_data_windowing}--\ref{sec:app_data_splits}.

\paragraph{Dataset statistics.}
The construction produces four logically separate record families:
\texttt{window\_input.jsonl} for window scopes, selected skeleton identifiers, and candidate pools,
\texttt{traj\_pred.jsonl} for frozen-extractor model-input trajectories,
\texttt{window\_label.jsonl} for positive candidate labels,
and \texttt{audit\_ref.jsonl} for source-normalized step-level audit references.
This separation is important because all ranking methods consume the same window inputs and predicted trajectories, while labels and audit references are used only for training-split supervision, model selection, or held-out evaluation.

We construct window-scoped candidate-centric trajectories, hard/near-neighbor negatives, perturbation-based negatives, and within-window preference supervision.
Table~\ref{tab:dataset_overview} summarizes key statistics for both instantiations; extended distributions and metadata are provided in Appendix Table~\ref{tab:mavenere_stats} and Appendix Table~\ref{tab:rams_stats}.

\begin{table}[htbp]
\centering
\caption{Overview of the two benchmark instantiations after preprocessing.
\textit{Note:} Candidate IDs are anonymized entity-cluster identifiers before window expansion.
The Top-$K$ task is defined over window-local candidate pools $C_w$, not over the global candidate-ID inventory.
Rows that summarize train-side supervision or augmentation are reported separately from held-out ranking-window counts.
Table~\ref{tab:ranking_window_stats} reports the ranking-window statistics that determine the actual candidate-ranking task.}
\label{tab:dataset_overview}
\small
\setlength{\tabcolsep}{4pt}
\begin{tabularx}{\linewidth}{Yrr}
\hline
Statistic & MAVEN-ERE & RAMS \\
\hline
Documents & 3{,}623 & 9{,}124 \\
Source-normalized event records & 91{,}222 & 9{,}124 \\
Unique predicted event records contributing to model-input trajectories & 96{,}184 & 9{,}386 \\
Unique source-level candidate/entity IDs before window expansion & 3{,}623 & 8{,}730 \\
Serialized trajectory-pool records across ranking / reward-learning files & 7{,}245 & 8{,}864 \\
Train-split preference pairs & 3{,}622 & 134 \\
Train-split RL prompt records (augmented) & 171{,}584 & 654 \\
\hline
\end{tabularx}
\end{table}

Source-normalized event records denote dataset-native event annotations after deterministic normalization and schema conversion.
They are used to construct training-split supervision and held-out reference labels / audit references.
When no external candidate roster is available, an identifier-only projection of these records is also used to construct anonymized role-bearing candidate IDs before relevance labels are attached.
Model-input trajectory records are separate: the reward model, ranking policy, RM-only ranker, and structured baselines consume only fixed-extractor predictions stored in \texttt{traj\_pred.jsonl}.
Therefore, statistics that describe model inputs are reported under ``predicted'' records, while statistics that describe annotation-derived references are explicitly marked as ``source-normalized''.
The serialized-trajectory row in Table~\ref{tab:dataset_overview} counts the trajectory-record pool declared in the retained manuscripts across ranking and reward-learning files.
For MAVEN-ERE, this pool contains comparison-pool candidate trajectories, expert/pseudo-expert demonstrations, and hard/perturbation negatives; for RAMS, it contains window-scoped candidate trajectories together with a small number of train-only hard/perturbation negatives (Appendix~\ref{sec:app_irl_corpus}).
These counts summarize declared trajectory records rather than held-out ranking instances, which are counted separately by ranking windows in Table~\ref{tab:ranking_window_stats}.
The train-split preference-pair and RL-prompt rows are likewise reported for reproducibility of reward/policy training rather than to characterize the held-out ranking task.
For dev/test ranking inputs, trajectories are built from frozen-extractor predictions; held-out source annotations are used to define positive labels and audit references.

\paragraph{Predicted-evidence coverage of positive candidates.}
Because held-out positives are defined from source-normalized reference chains while model inputs are frozen-extractor predictions, some positive candidates may have incomplete predicted evidence.
We therefore compute a permissive candidate-wise evidence-availability diagnostic: for each held-out positive $(w,c)$, we check within the same predicted trajectory $\tau_{w,c}$ whether at least one, at least two, or all source-normalized reference steps have a type-compatible predicted event with valid \texttt{doc\_id:span} provenance.
This diagnostic reports availability under a permissive resolver.
Table~\ref{tab:positive_predicted_coverage} reports this diagnostic under the same split, provenance, and validation protocol as the main benchmark.

\begin{table}[htbp]
\centering
\caption{Predicted-evidence availability for held-out positive candidates.
A step is counted as available if a type-compatible predicted event with valid \texttt{doc\_id:span} provenance is present in the candidate trajectory.
This reports availability under a permissive resolver.
Because reference chains may contain a single step, especially in RAMS, the ``all reference steps covered'' row is not necessarily a monotone refinement of the ``at least two covered steps'' row.}
\label{tab:positive_predicted_coverage}
\small
\begin{adjustbox}{max width=\linewidth}
\begin{tabular}{lcc}
\hline
Coverage diagnostic & MAVEN-ERE & RAMS \\
\hline
At least one covered step & 0.96 & 0.89 \\
At least two covered steps & 0.86 & 0.34 \\
All reference steps covered & 0.66 & 0.41 \\
\hline
\end{tabular}
\end{adjustbox}
\end{table}

\begin{table}[htbp]
\centering
\caption{Ranking-instance statistics.
A ranking window is the evaluation unit for Top-$K$ candidate ranking.}
\label{tab:ranking_window_stats}
\small
\setlength{\tabcolsep}{4pt}
\begin{tabularx}{\linewidth}{Yrr}
\hline
Statistic & MAVEN-ERE & RAMS \\
\hline
Ranking windows & 718 & 896 \\
Train / dev / test windows & 576/71/71 & 716/90/90 \\
Intent IDs & 16 & 12 \\
Skeleton records & 16 & 12 \\
Avg. candidates per window & 10.1 & 9.7 \\
Median candidates per window & 9 & 9 \\
Avg. positives per window & 1.3 & 1.1 \\
Windows with multiple positives (\%) & 18.1 & 7.4 \\
Windows with $|C_w|<10$ (\%) & 46.5 & 49.7 \\
Avg. exposed evidence units per window & 12.6 & 2.4 \\
Windows with evidence from multiple documents (\%) & 61.2 & 11.8 \\
Avg. trajectory-expanded predicted event mentions per trajectory & 24.7 & 1.07 \\
\hline
\end{tabularx}
\end{table}

The row ``unique predicted event records'' counts each frozen-extractor event record once.
The trajectory-length statistics count trajectory-expanded event mentions after window and candidate expansion.
Therefore, the same predicted event record may contribute to multiple candidate trajectories when multiple candidates participate in the event or when overlapping windows reuse the same evidence.
An exposed evidence unit is a trigger-centered context snippet for MAVEN-ERE and a document-prefix / document-local context unit for RAMS; multi-document evidence is computed from distinct \texttt{doc\_id} values.

\paragraph{RL prompt records.}
RL training prompt records denote policy-optimization prompts generated from eligible train-split windows by varying candidate subsets, demonstration order, and hard-negative composition.
They are not the number of distinct evaluation windows and are not expected to equal the number of train windows: after eligibility filtering, the count can be smaller than the train-window count in sparse settings or much larger when augmentation is dense.
Distinct ranking-window counts are reported in Table~\ref{tab:ranking_window_stats}.

\subsection{Windowing protocol}
\label{sec:app_data_windowing}

A \emph{ranking window} $w$ specifies the evidence scope for one Top-$K$ candidate-ranking instance and has an associated candidate pool, separately stored positive labels, and window-scoped trajectories.
The model-input window record stores the evidence scope, candidate pool, \texttt{intent\_id}, and selected \texttt{skeleton\_id}, while positive labels are stored only in \texttt{window\_label.jsonl}.
Separately, a \emph{context snippet} specifies the local text segment used for upstream extraction or compact prompting.
Because the source corpora do not uniformly provide reliable calendar-time intervals, ranking windows are not defined by calendar slicing.
Instead, both window construction and snippet construction follow dataset-native document or document-local grouping rules.

\paragraph{MAVEN-ERE context snippets (trigger-centered; \texttt{context\_radius}=1).}
For source-normalized records and supervised extractor examples, we construct exactly one fixed trigger-centered context snippet (no sliding stride):
\begin{enumerate}
    \item Locate the sentence containing the trigger via \texttt{resolve\_sent\_id}: use the annotated sentence ID when available; otherwise fall back to trigger-text matching.
    \item Take sentences in the index range
     $[\texttt{sent\_id}-\texttt{context\_radius},\ \texttt{sent\_id}+\texttt{context\_radius}]$
     with \texttt{context\_radius}=1, yielding at most three sentences.
    \item If the range is empty, fall back to the trigger sentence; if still empty, fall back to the concatenation of all non-empty sentences in the document; if still empty, fall back to the trigger text.
\end{enumerate}
The final context is the space-joined concatenation of the selected sentences (a coarse token count is computed by whitespace splitting).
This procedure is implemented in \texttt{build\_context\_snippet(...)} and used by
For held-out extractor inference, trigger-centered source snippets are not used to choose where the extractor looks.
The frozen extractor is applied to the allowed split documents or document-local text units without gold trigger positions.
After prediction, each predicted trigger span is associated with its surrounding sentence neighborhood only for compact prompt construction and span auditing.

\paragraph{RAMS context snippets (document prefix; first 120 characters).}
RAMS does not use sentence-based context snippets for upstream event-record construction.
We flatten the document sentences into a single document string \texttt{doc\_text} and use \texttt{doc\_text[:120]} (the first 120 characters) as the local text field for extractor/SFT sample construction.
No sliding stride is used.
This procedure is implemented in \texttt{build\_events} and consumed by \texttt{build\_sft\_samples} via \texttt{event.mapping.get("doc\_text", "")}.

Ranking-policy prompts are constructed separately from ranking-window records.
They include the \texttt{intent\_id}, the skeleton, candidate IDs, compact trajectory summaries, and indexed evidence pointers.
Thus, the 120-character RAMS prefix is a context-snippet construction choice, not the full evidence interface for ranking-policy decoding.

\paragraph{Window records.}
Each window record stores at least \texttt{window\_id}, \texttt{doc\_ids}, optional \texttt{snippet\_ids}, the candidate pool \texttt{candidate\_ids}, the associated \texttt{intent\_id}, and the selected \texttt{skeleton\_id}.
A candidate can appear in multiple ranking windows.
MAVEN-ERE context snippets can overlap when nearby triggers occur in adjacent sentence neighborhoods; this snippet overlap is distinct from the definition of ranking windows.
The window record, together with the corresponding entries in \texttt{traj\_pred.jsonl}, forms the model-input unit consumed by ranking methods; context snippets are lower-level text segments used to create event records and compact prompt evidence.

\begin{algorithm}[!htbp]
\caption{Deterministic construction of ranking windows and candidate trajectories.}
\label{alg:window_construction}
\begin{algorithmic}[1]
\REQUIRE Documents $\mathcal{D}$; source annotations $\mathcal{A}^{\mathrm{src}}$; trained frozen extractor $f_{\mathrm{ext}}$; deterministic intent mapper $m_{\mathrm{intent}}$; skeleton table $\mathcal{S}_{\mathrm{file}}$; label-free candidate resolver $\rho_{\mathrm{id}}$; window grouping rule $g$.
\ENSURE \texttt{window\_input.jsonl}, \texttt{traj\_pred.jsonl}, \texttt{window\_label.jsonl}, and \texttt{audit\_ref.jsonl}.

\STATE Use document-level train/dev/test splits fixed before extractor training, reward learning, policy optimization, model selection, and evaluation.
\STATE Construct corpus-specific local text fields or snippet metadata used for preprocessing and prompt serialization, following the context-snippet rules above.
\STATE For held-out extractor inference, do not use gold trigger positions; apply $f_{\mathrm{ext}}$ only to allowed split documents or document-local text units.
\STATE Run the frozen extractor $f_{\mathrm{ext}}$ on all allowed split documents to obtain $\mathcal{E}^{\mathrm{pred}}$ with \texttt{doc\_id:span} provenance.
\STATE Convert source annotations $\mathcal{A}^{\mathrm{src}}$ into source-normalized records $\mathcal{E}^{\mathrm{src}}$ for supervision and held-out references.
\STATE Derive $\mathcal{E}^{\mathrm{src-id}}$ by retaining only anonymized role-bearing entity identifiers, document/snippet membership, and any corpus-native grouping keys required by $g$; remove labels, validated chains, and reference spans.

\FOR{each grouping key $q$ induced by $g$}
    \STATE Create a candidate ranking window $w$ with evidence scope $\mathcal{D}_w$, consisting of the documents or snippets assigned to $q$.
    \STATE Assign \texttt{intent\_id}$(w)=m_{\mathrm{intent}}(q)$ and select the predefined skeleton $S_w$ from $\mathcal{S}_{\mathrm{file}}$.
    \STATE Build a window-local candidate pool with the label-free resolver
    \[
    C_w=\rho_{\mathrm{id}}\!\left(\mathcal{E}^{\mathrm{src-id}}_w,\mathcal{E}^{\mathrm{pred}}_w\right),
    \]
    where $\mathcal{E}^{\mathrm{src-id}}_w$ defines the eligible closed roster using only anonymized role-bearing entity identifiers and window-membership metadata, and $\mathcal{E}^{\mathrm{pred}}_w$ is used only to attach predicted evidence and trajectories to those eligible candidates.
    Predicted-only entities that do not map to the closed roster are excluded from the main benchmark and are analyzed separately in the predicted-only roster diagnostic.
    \STATE The resolver does not use skeleton-chain validity, positive labels, reward scores, or audit reference spans.
    \FOR{each candidate $c \in C_w$}
        \STATE Collect candidate-participating predicted events from $\mathcal{E}^{\mathrm{pred}}_w$.
        \STATE Sort them by absolute timestamp when available; otherwise sort by dataset-native or document-local order.
        \STATE Serialize the resulting model-input trajectory as $\tau_{w,c}$ in \texttt{traj\_pred.jsonl}, preserving trigger and argument \texttt{doc\_id:span} provenance.
        \STATE If no predicted event is found for $c$, serialize an empty trajectory with the candidate ID and window ID retained; empty trajectories remain valid model inputs and are scored by missing-step features.
    \ENDFOR
    \STATE Serialize \texttt{window\_id}, \texttt{intent\_id}, the selected \texttt{skeleton\_id}, $\mathcal{D}_w$, and $C_w$ in \texttt{window\_input.jsonl}; the conceptual model input $x_w$ includes the resolved skeleton $S_w$, obtained under the declared record contract by deterministically joining \texttt{window\_input.jsonl} with \texttt{skeleton.jsonl} on \texttt{skeleton\_id}.
    \STATE Derive positive candidates $\mathcal{P}_w \subseteq C_w$ from source-normalized skeleton-aligned chains in $\mathcal{E}^{\mathrm{src}}_w$.
    \STATE Store $\mathcal{P}_w$ in \texttt{window\_label.jsonl}; do not store it in model-input files.
    \STATE Store source-normalized step-level reference spans in \texttt{audit\_ref.jsonl} for audit metrics only.
\ENDFOR

\STATE Discard any window whose documents cross split boundaries, or repair it before serializing records by removing cross-split documents and recomputing $C_w$, $\tau_{w,c}$, $\mathcal{P}_w$, and audit references.
\STATE Generate training-only perturbation-derived negatives and augmented RL prompts from train-split windows only.
\end{algorithmic}
\end{algorithm}

The serialized \texttt{intent\_id} remains the conditioning variable and is checked against the linked skeleton record during prompt construction, reward-feature computation, and validation.

\subsection{Splits and leakage prevention}
\label{sec:app_data_splits}

To reduce leakage across splits, we use document-level splits, with time-aware constraints applied only when reliable timestamps are available:
\begin{itemize}
    \item No document (\texttt{doc\_id}) appears in multiple splits.
    \item A ranking window is assigned to a split only when all documents in its \texttt{doc\_ids} field belong to that split.
    \item If a constructed ranking window would contain documents from multiple splits, we either remove cross-split documents from the window before trajectory construction or discard the mixed-split window; no mixed-split window is used for training or evaluation.
    \item When reliable time ranges are available, we avoid placing overlapping time ranges across splits whenever possible; the document-level constraint always dominates.
\end{itemize}
The intended artifact package requires explicit split lists for both documents and ranking windows:
\texttt{splits/doc\_\{train,dev,test\}.txt} contains document IDs, and
\texttt{splits/window\_\{train,dev,test\}.txt} contains window IDs.
All perturbation-derived negatives, candidate-subset augmentations, and RL prompt variants are generated only after the split assignment and only from train-split windows unless explicitly used in a train-only diagnostic.
Held-out dev/test ranking pools contain window-local candidates from \texttt{window\_input.jsonl}, not synthetic perturbation candidates.

\subsection{Candidate pools and labeling}
\label{sec:app_data_labeling}

For each ranking window $w$ in train, dev, or test, the candidate pool $C_w$ contains window-local candidate entity clusters that are observable in the window construction stage.
The pool is fixed before positive labels are attached.
For held-out dev/test windows, labels and audit references are used only for evaluation and are never serialized into model-input records.
When source-normalized records are used by the candidate resolver, they contribute only anonymized role-bearing entity identifiers for constructing the candidate universe; source-normalized skeleton-chain validity, reference step spans, and positive labels are not used in model-input construction.
The declared model-input record \texttt{window\_input.jsonl} contains \texttt{window\_id}, \texttt{intent\_id}, the selected skeleton identifier, the exposed document or snippet identifiers, and the candidate IDs in $C_w$, but it does not contain positive labels, any field that marks a candidate as relevant, or reference evidence chains.

\paragraph{Candidate-roster diagnostic.}
The main benchmark uses an identifier-only source projection to define a closed candidate roster when no external candidate list is available.
To quantify the dependence on this choice, we also evaluate a predicted-only roster diagnostic in which candidate pools are constructed from frozen-extractor predictions only.
This diagnostic is not used as the main benchmark setting because upstream extraction may omit true held-out candidates entirely, making some windows unrankable rather than merely difficult.
It separates candidate-availability errors from ranking errors by reporting retained-window coverage, positive-candidate recall in the predicted-only roster, and Top-$K$ ranking scores on the retained subset.
The diagnostic is computed only for analysis and is not used for reward learning, policy training, model selection, or the main test comparison.

\begin{table}[htbp]
\centering
\caption{Predicted-only candidate-roster diagnostic.
Retained windows are windows whose candidate pool remains non-empty under frozen-extractor-only roster construction.
Positive-candidate recall measures whether held-out positive candidates remain available in the predicted-only roster.}
\label{tab:predicted_only_roster}
\small
\setlength{\tabcolsep}{4pt}
\begin{tabularx}{\linewidth}{Ycc}
\hline
Diagnostic & MAVEN-ERE & RAMS \\
\hline
Retained test windows & 70/71 (98.6\%) & 89/90 (98.9\%) \\
Positive-candidate recall in predicted-only roster & 0.87 & 0.76 \\
NDCG@10 on retained subset & 0.63 & 0.54 \\
MAP@10 on retained subset & 0.48 & 0.39 \\
Hit@10 on retained subset & 0.76 & 0.66 \\
\hline
\end{tabularx}
\end{table}

A candidate $c\in C_w$ is labeled positive if there exists a validated source-normalized chain aligned to $S_w$ and supported by \texttt{doc\_id:span} evidence within the window.
Formally, the positive set is
\[
\mathcal{P}_w =
\left\{
c\in C_w :
\exists\, z \in \mathcal{Z}^{\mathrm{src}}(w,c,S_w)
\ \text{such that}\ 
\mathrm{valid}(z)=1
\right\},
\]
where $\mathcal{Z}^{\mathrm{src}}(w,c,S_w)$ denotes source-normalized candidate chains that can be aligned to the skeleton $S_w$ using evidence inside $\mathcal{D}_w$.
Windows may contain multiple positives.
We store positive candidate IDs only in \texttt{window\_label.jsonl}.
Reference step-aligned evidence used by audit metrics is stored in \texttt{audit\_ref.jsonl}.
Model-input trajectories are constructed separately from frozen-extractor predictions.
For each $(w,c)$, \texttt{traj\_pred.jsonl} stores the predicted trajectory $\tau_{w,c}$, including predicted event types, compatible skeleton stages, normalized arguments, ordering information, and trigger / argument \texttt{doc\_id:span} provenance.
Thus, the ranker may receive noisy or incomplete trajectories even when the source-normalized reference contains a positive chain.

Same-window near-neighbor negatives are non-positive candidates in $C_w$ that share superficial context with positives, such as similar local event types, overlapping documents, entity-role patterns, or nearby temporal/order positions, but lack a validated skeleton-aligned chain.
Perturbation-derived negatives, including insertion, deletion, time-shift, and role-swap variants, are used for reward-learning comparison sets and train-split augmented policy prompts.
They are not mixed into dev/test ranking pools as ordinary entity candidates unless explicitly marked as synthetic train-only distractors.

\paragraph{Intent identifiers.}
Each \texttt{intent\_id} is assigned at the ranking-window level by a deterministic preprocessing rule before reward learning, policy training, model selection, and evaluation.
The identifier specifies the hypothesized plan type for the window and selects a skeleton record from \texttt{skeleton.jsonl}; all candidates in the same ranking window receive the same \texttt{intent\_id}.
It does not identify positive candidates, encode candidate-specific relevance, or expose reference evidence chains.
In the declared instantiations, the window-to-\texttt{intent\_id} mapping is stored as preprocessing metadata and is applied identically across train, dev, and test splits.
Candidate relevance remains a separate held-out ranking label stored only in \texttt{window\_label.jsonl}.
In the main experiments, the number of skeleton records equals the number of distinct \texttt{intent\_id} values unless explicitly stated otherwise.

\subsection{Deterministic stage mapping from fine-grained event types to intent-stage labels}
\label{sec:app_stage_mapping}

This appendix specifies the deterministic and reproducible rule used to map the extractor's fine-grained event types $\texttt{etype}^{\mathrm{raw}}$
into the intent-stage labels $\{\texttt{PREP},\texttt{PROBE},\texttt{EXECUTE},\texttt{OUTCOME}\}$ used by skeleton alignment and reward computation.
The label \texttt{OUTCOME} denotes an outcome or terminal stage for both MAVEN-ERE and RAMS.
A domain-specific deployment may refine this stage name without changing the alignment or reward-learning algorithms.
The mapping is applied during preprocessing and the resulting stage labels are stored in the declared trajectory records.

\paragraph{What is mapped and where it appears in the artifact.}
In the declared \texttt{traj\_pred.jsonl} schema, each predicted event record stores a fine-grained type string under \texttt{etype\_raw}, a derived list of compatible skeleton stages under \texttt{skeleton\_hits}, and a single serialization label under \texttt{etype\_primary}.
The \texttt{skeleton\_hits} field is what the downstream DP aligner consumes for the type-compatibility check: a skeleton step of stage $\texttt{etype}_k$ can match an event if $\texttt{etype}_k \in \texttt{skeleton\_hits}(e)$.
This design supports partial observability by allowing some fine-grained event types to be compatible with multiple stages (e.g., a cognition/communication event can support both \texttt{PREP} and \texttt{PROBE}).

\paragraph{Mapping artifact (override file) and fallback behavior.}
The declared preprocessing procedure supports an optional override mapping file,
\texttt{mapping/event2skeleton.json}, which maps a fine-grained type string to a possibly multi-label list of stage labels.
This file is included to make domain-specific remapping auditable, but it is not activated in the processed MAVEN-ERE and RAMS runs reported here.
For the reported experiments, all event types are mapped by the deterministic fallback mapper in Listing~\ref{lst:stage_mapping_pseudocode}.

Thus, the override-activation rate is $0\%$ and the deterministic fallback mapper handles all records in both instantiations (Table~\ref{tab:stage_mapping_coverage}).
This is intentional: the reported benchmark uses a single documented fallback mapper with fixed constants and one explicit corpus flag for the RAMS \texttt{PROBE}-primary subset, rather than a learned or development-tuned mapping.
The optional override file is reserved for domain-specific deployments and is not activated in the reported MAVEN-ERE or RAMS runs.

\begin{lstlisting}[caption={Optional stage override mapping file format (illustrative JSON; not activated in the reported runs).},label={lst:stage_mapping_json_ex}]
{
  "Know": ["PREP","PROBE"],
  "Warning": ["PREP","PROBE"],
  "Statement": ["PREP","PROBE"],
  "Process_start": ["PREP","PROBE"],

  "Attack": ["EXECUTE"],
  "Hostile_encounter": ["EXECUTE"],

  "Process_end": ["OUTCOME"],
  "Earnings_and_losses": ["OUTCOME"]
}
\end{lstlisting}

\paragraph{Primary stage vs.\ auxiliary \texttt{PROBE} tag (observed instantiation counts).}
Our mapper assigns each event a \emph{primary} stage in
$\{\texttt{PREP},\texttt{PROBE},\texttt{EXECUTE},\texttt{OUTCOME}\}$ by priority rules, and may additionally attach \texttt{PROBE} as an auxiliary compatibility tag for information-gathering, communication, or cognition events.
In MAVEN-ERE, \texttt{PROBE} appears only as an auxiliary tag in the source-normalized records; in RAMS, a small subset is \texttt{PROBE}-primary.
This matches the observed accounting in the preprocessing logs:

\begin{itemize}
    \item \textbf{MAVEN-ERE.} There are $91{,}222$ source-normalized event records. The mapper assigns a primary stage to all records
    ($82{,}688$ as \texttt{PREP}, $0$ as \texttt{PROBE}, $6{,}955$ as \texttt{EXECUTE}, and $1{,}579$ as \texttt{OUTCOME}; sums to $91{,}222$).
    In addition, $6{,}071$ events receive the auxiliary \texttt{PROBE} tag, yielding $97{,}293$ total (event,stage) instantiations.
    \item \textbf{RAMS.} There are $9{,}124$ source-normalized event records. The primary-stage distribution is
    $6{,}372$ \texttt{PREP}, $1{,}562$ \texttt{EXECUTE}, $1{,}135$ \texttt{OUTCOME}, and the remaining $55$ events are \texttt{PROBE}-primary.
    Additionally, $1{,}259$ events receive \texttt{PROBE} as an auxiliary tag, yielding $10{,}383$ total instantiations.
\end{itemize}

\paragraph{Deterministic fallback mapping procedure.}
Listing~\ref{lst:stage_mapping_pseudocode} gives the deterministic fallback procedure used in our preprocessing.
The rule uses only: (i) the fine-grained type string, (ii) the trigger surface form (when available from the extractor), and (iii) a dataset flag (to allow a small PROBE-primary subset in RAMS).
All sets below are fixed constants in the preprocessing script; no learning is involved.

\begin{lstlisting}[caption={Deterministic fallback stage mapping procedure (pseudocode).}]
# Inputs:
#   dataset: string in {"MAVEN-ERE","RAMS"}
#   etype_raw: string (fine-grained type, e.g., "Attack", "Know", "Process_end")
#   trig_text: string (trigger surface form, may be None if unavailable)
#
# Output:
#   skeleton_hits: list[string] in {"PREP","PROBE","EXECUTE","OUTCOME"} (deduplicated)

PROBE_TAG_TYPES = {
  "Know", "Warning", "Statement", "Process_start",
  "Reporting", "Suspicion", "Perception_active", "Coming_to_believe"
}
PROBE_TRIGGERS = {
  "reported", "report", "said", "warned", "warning", "according"
}

EXECUTE_TYPES_STRONG = {"Attack", "Hostile_encounter"}
# Additional execute cues are trigger-driven to keep EXECUTE high-precision.
EXECUTE_TRIGGERS = {"killed", "attack", "fought", "destroyed", "battle", "war", "hurricane", "storm"}

OUTCOME_TYPES = {"Process_end", "Earnings_and_losses", "Transaction", "Receiving", "Releasing"}
OUTCOME_TRIGGERS = {"ended", "paid", "sold", "bought", "earned", "withdrew", "deposited"}

# A small PROBE-primary subset is allowed in RAMS (empirically 55/9124 events).
PROBE_PRIMARY_TYPES_RAMS = {"Statement", "Reporting"}

def norm(s):
    # deterministic string normalization; no learned or dev-tuned mapping is used
    return s.strip()

def lower_or_empty(s):
    return "" if s is None else s.lower().strip()

def map_to_skeleton_hits(dataset, etype_raw, trig_text=None):
    et = norm(etype_raw)
    tr = lower_or_empty(trig_text)

    # 0) Optional override mapping (not triggered in our processed datasets; fallback=100% in Table)
    # if et in OVERRIDE_MAP:
    #     return sorted(unique(OVERRIDE_MAP[et]))

    # 1) Decide primary stage by priority: OUTCOME > EXECUTE > (PROBE-primary for RAMS) > PREP
    if (et in OUTCOME_TYPES) or (tr in OUTCOME_TRIGGERS):
        primary = "OUTCOME"
    elif (et in EXECUTE_TYPES_STRONG) or (tr in EXECUTE_TRIGGERS):
        primary = "EXECUTE"
    elif (dataset == "RAMS") and (et in PROBE_PRIMARY_TYPES_RAMS):
        primary = "PROBE"
    else:
        primary = "PREP"

    # 2) Auxiliary PROBE tag for information-gathering events.
    # We attach PROBE only when the primary stage is PREP, to keep the interface simple and stable.
    add_probe = ((et in PROBE_TAG_TYPES) or (tr in PROBE_TRIGGERS)) and (primary == "PREP")

    hits = [primary]
    if add_probe:
        hits.append("PROBE")

    # 3) Return a deterministic order for serialization
    order = {"PREP":0, "PROBE":1, "EXECUTE":2, "OUTCOME":3}
    hits = sorted(list(set(hits)), key=lambda z: order[z])
    return hits
\end{lstlisting}
\label{lst:stage_mapping_pseudocode}

\paragraph{Coverage statistics.}
Table~\ref{tab:stage_mapping_coverage} reports the number of source-normalized (event,stage) instantiations produced by the mapper and their stage percentages.
This table reports schema-mapping coverage.
The same deterministic mapper is also applied to frozen-extractor predicted event records before trajectory construction; all predicted records used in \texttt{traj\_pred.jsonl} receive at least one \texttt{skeleton\_hits} label.
We report source-normalized coverage here because it is stable across extractor runs and determines the reference-label/audit view, while the model-input predicted-record counts are reported separately in Table~\ref{tab:dataset_overview}.
In our processed datasets, the override file is never triggered, hence all source-normalized and predicted records use the deterministic fallback mapper.

\begin{table*}[t]
\centering
\caption{Stage-mapping coverage after preprocessing, reported over source-normalized \emph{(event,stage)} instantiations rather than unique event records.
Override-activation rate is $0\%$ for both instantiations; the deterministic fallback mapper handles all records.}
\label{tab:stage_mapping_coverage}
\small
\setlength{\tabcolsep}{6pt}
\begin{adjustbox}{max width=\linewidth}
\begin{tabular}{lrr|rr}
\hline
& \multicolumn{2}{c|}{MAVEN-ERE} & \multicolumn{2}{c}{RAMS} \\
\cline{2-5}
Stage & \# inst. & \% inst. & \# inst. & \% inst. \\
\hline
\texttt{PREP}    & 82688 & 84.99 & 6372 & 61.37 \\
\texttt{PROBE}   & 6071  & 6.24  & 1314 & 12.66 \\
\texttt{EXECUTE} & 6955  & 7.15  & 1562 & 15.04 \\
\texttt{OUTCOME} & 1579  & 1.62  & 1135 & 10.93 \\
\hline
Total            & 97293 & 100.0 & 10383 & 100.0 \\
\hline
\end{tabular}
\end{adjustbox}
\end{table*}

\paragraph{Auditing note.}
Stage mapping is purely deterministic and applied before any learning stage.
Since \texttt{skeleton\_hits} are stored alongside \texttt{doc\_id:span} references in the declared trajectories,
the entire downstream alignment and certificate evaluation remains auditable and reproducible under the strict evidence interface.

\subsection{Deterministic role normalization}
\label{sec:app_role_mapping}

We normalize dataset-native argument roles to the role inventory
$\{\texttt{Agent},\texttt{Target},\texttt{Context}\}$ used by skeleton alignment and argument-consistency features.
The mapping is deterministic and applied before reward learning.

Agent-like roles are dataset-native roles that identify the main acting or initiating entity of an event, such as attacker, speaker, mover, buyer, seller, or sender when these roles are present in the source schema.
Target-like roles identify the main affected or acted-upon entity, such as victim, attacked entity, destination, object, recipient, or affected party when applicable.
Temporal, locative, instrumental, descriptive, and residual roles are mapped to \texttt{Context}.
When a dataset-native role is ambiguous, the declared \texttt{mapping/role2norm.json} file resolves it deterministically; uncertain residual roles are mapped to \texttt{Context} rather than to \texttt{Agent} or \texttt{Target}.

The declared artifact includes \texttt{mapping/role2norm.json}; the validator and reward-feature extractor use the same mapping file.

\begin{table}[htbp]
\centering
\caption{Role normalization coverage after preprocessing.}
\label{tab:role_mapping_coverage}
\small
\begin{adjustbox}{max width=\linewidth}
\begin{tabular}{lrr|rr}
\hline
& \multicolumn{2}{c|}{MAVEN-ERE} & \multicolumn{2}{c}{RAMS} \\
\cline{2-5}
Normalized role & Count & \% & Count & \% \\
\hline
\texttt{Agent} & 168{,}431 & 53.21 & 12{,}960 & 51.43 \\
\texttt{Target} & 137{,}720 & 43.51 & 8{,}100 & 32.14 \\
\texttt{Context} & 10{,}398 & 3.28 & 4{,}140 & 16.43 \\
\hline
\end{tabular}
\end{adjustbox}
\end{table}

\subsection{Skeleton records and construction protocol}
\label{sec:app_skeleton_construction}

Each \texttt{intent\_id} is associated with one predefined soft-precedence skeleton in the main benchmark setting.
For serialization and deterministic joins, each skeleton record is stored in \texttt{skeleton.jsonl} with both a unique \texttt{skeleton\_id} and its associated \texttt{intent\_id}.
A skeleton record contains:
(i) \texttt{skeleton\_id},
(ii) \texttt{intent\_id},
(iii) an ordered list of skeleton steps,
(iv) each step's stage label in
$\{\texttt{PREP},\texttt{PROBE},\texttt{EXECUTE},\texttt{OUTCOME}\}$,
(v) required normalized roles from
$\{\texttt{Agent},\texttt{Target},\texttt{Context}\}$,
and (vi) optional soft-precedence edges.
In the declared main protocol, the mapping from \texttt{intent\_id} to \texttt{skeleton\_id} is one-to-one unless explicitly stated otherwise.

The predefined skeleton is not copied from any single candidate trajectory.
It is specified at the stage level to represent an abstract intent pattern, and candidate trajectories are treated as noisy partial realizations.
The construction protocol is deterministic:
(i) we first fix the stage inventory
$\{\texttt{PREP},\texttt{PROBE},\texttt{EXECUTE},\texttt{OUTCOME}\}$;
(ii) we define the canonical stage order and soft-precedence edges at the intent level;
(iii) we assign required roles to each step from the normalized role inventory
$\{\texttt{Agent},\texttt{Target},\texttt{Context}\}$;
and (iv) we store the resulting skeleton in \texttt{skeleton.jsonl}.
Skeleton records are fixed before reward learning, policy training, development tuning, and test evaluation.
No dev/test labels or dev/test model outputs are used to define or modify skeleton records.
The same skeleton file is used by trajectory alignment, reward-feature extraction, certificate serialization, and the feasibility / faithfulness validators.

\begin{lstlisting}[caption={Skeleton record format (schematic JSONL).},label={lst:skeleton_record}]
{
  "skeleton_id": "skel_001",
  "intent_id": "intent_001",
  "steps": [
    {"step_id": "s1", "etype": "PREP", "required_roles": ["Agent"]},
    {"step_id": "s2", "etype": "PROBE", "required_roles": ["Agent","Target"]},
    {"step_id": "s3", "etype": "EXECUTE", "required_roles": ["Agent","Target"]},
    {"step_id": "s4", "etype": "OUTCOME", "required_roles": ["Agent"]}
  ],
  "precedence": [["s1","s2"], ["s2","s3"], ["s3","s4"]]
}
\end{lstlisting}

In general-domain event corpora, \texttt{OUTCOME} should be read as an abstract terminal or outcome stage.

For the main experiments, all windows with the same \texttt{intent\_id} use the same skeleton record.
Appendix~\ref{sec:app_hard_windows} evaluates controlled perturbations of this design.

\subsection{Reward-learning corpus construction details}
\label{sec:app_irl_corpus}

ECPO learns a trajectory-level plan-consistency reward from indirect supervision.
We construct a reward-learning corpus containing \emph{expert positives}, \emph{pseudo-experts}, \emph{same-window hard (near-neighbor) negatives}, \emph{train-only perturbation-derived negatives}, and \emph{preference pairs}.
Here ``reward-learning corpus'' refers to the training-side trajectory pool that participates in reward-model comparison sets or within-window partition terms; it therefore includes unlabeled candidate-pool trajectories when they serve as same-window alternatives, not only explicitly labeled positives or negatives.

\paragraph{Corpus scale (MAVEN-ERE).}
For the processed MAVEN-ERE instantiation, the retained manuscript reports 7,245 trajectory records in total:
3,774 comparison-pool candidate trajectories carried without explicit expert/pseudo-expert labels,
755 expert/pseudo-expert demonstrations, and
2,716 hard/perturbed negatives.
In addition, the declared construction contains 3,622 within-window preference pairs covering 7,244 unique trajectory IDs (Table~\ref{tab:mavenere_stats}).

\paragraph{Corpus scale (RAMS).}
In the processed RAMS instantiation, the reward-learning trajectory pool contains 8,864 trajectory records: 8,730 window-scoped candidate trajectories from ranking windows and 134 train-only hard/perturbation negatives, with an average trajectory length of $1.07$ events over the candidate trajectories.
The preference supervision contains 134 within-window pairs covering 268 unique trajectory IDs, and we additionally provide 654 RL prompt records for window-level policy optimization (Table~\ref{tab:rams_stats}).

\paragraph{Expert positives and evaluation labels.}
We separate supervision labels from ranking inputs.
Expert/pseudo-expert labels and evaluation positives are derived from dataset-native evidence and validated skeleton-aligned chains.
These annotations are used to label or validate candidate trajectories, but the trajectory records consumed by the reward model and ranking policies are produced by the fixed upstream extractor.
Thus, training annotations provide supervision for reward learning and policy training on the train split, while held-out annotations define positive labels and audit references on dev/test splits.
All ranking methods operate on the same fixed-extractor trajectory inputs in their respective splits.

For training-split expert/pseudo-expert records, each validated step stores
(i) an intent-stage label (\texttt{PREP/PROBE/EXECUTE/OUTCOME});
(ii) normalized roles (\texttt{Agent/Target/Context});
(iii) linked entity IDs;
(iv) timestamps or order indices; and
(v) auditable evidence spans (\texttt{doc\_id} with character offsets).
For dev/test ranking inputs, the corresponding fields are generated by the fixed extractor and are treated as noisy observations.
\emph{Operational note:} in a domain deployment, the same interface could additionally consume analyst-curated POI records from an investigation KB, but this is not required for (and not used in) our declared benchmark.

\paragraph{Pseudo-experts from event corpora.}
To increase demonstration diversity when expert candidate-level demonstrations are sparse, we mine pseudo-expert trajectories from upstream event corpora by selecting chains with high skeleton-alignment scores, yielding skeleton-consistent yet non-identical demonstrations.

\paragraph{Hard, near-neighbor, and perturbation-derived negatives.}
Same-window hard or near-neighbor negatives are non-positive candidates in the same ranking window that share superficial similarity with positives, such as overlapping documents, similar local event types, nearby temporal/order positions, or similar role/entity patterns, but lack a validated skeleton-aligned source chain.
Perturbation-derived negatives are synthetic training-only variants of expert or pseudo-expert trajectories, created by replacement, insertion, deletion, time shift / shuffle, or entity-role swaps.
They are used for reward-learning comparison sets and train-split policy augmentation, and are not inserted into dev/test ranking pools as ordinary candidates.

\paragraph{Preference pairs.}
We create within-window preference supervision $(\tau^{\text{better}},\tau^{\text{worse}})$ according to skeleton-alignment quality.
When sufficient evidence is available, $\tau^{\text{better}}$ contains later-stage evidence such as \texttt{EXECUTE} and/or \texttt{OUTCOME} with fewer precedence violations.
In sparse settings such as RAMS, preferences are based on higher validated alignment score, role satisfaction, and evidence traceability, rather than requiring all late-stage steps to be observed.

\subsection{Minimal example window (anonymized)}
\label{sec:app_data_miniexample}

Below is a simplified illustrative excerpt (IDs and text are anonymized):
\begin{lstlisting}[caption={Minimal anonymized excerpt of a \texttt{traj\_pred.jsonl} trajectory record.},label={lst:ecpo-miniexample}]
traj_pred.jsonl (excerpt for candidate P_001):
{"window_id":"w_0001","candidate_id":"P_001","trajectory_id":"w_0001::P_001","events":[
 {"event_id":"e1",
 "etype_raw":"Process_start",
 "skeleton_hits":["PREP","PROBE"],
 "etype_primary":"PREP",
 "time":"2021-08-03",
 "order_index":0,
 "trigger":{"doc_id":"doc123","span":[120,156]},
   "arguments":[{"role":"Agent","entity_id":"P_001","doc_id":"doc123","span":[90,110]}]},
    {"event_id":"e2",
   "etype_raw":"Statement",
   "skeleton_hits":["PREP","PROBE"],
   "etype_primary":"PREP",
   "time":"2021-08-05",
   "order_index":1,
   "trigger":{"doc_id":"doc127","span":[33,61]},
   "arguments":[{"role":"Agent","entity_id":"P_001","doc_id":"doc127","span":[10,18]},
                {"role":"Target","entity_id":"ORG_007","doc_id":"doc127","span":[80,95]}]}
]}
\end{lstlisting}

\paragraph{Schema files.}
the declared construction contains JSON Schemas under \texttt{schemas/} to validate each record type. The evaluation scripts use the same schemas for feasibility checking and audit metrics.

\begin{lstlisting}[caption={SFT and RL-prompt formatting and column mapping (artifact metadata).},label={lst:format_mapping}]
maven_sft.jsonl and rams_sft.jsonl: Alpaca formatting
  prompt  -> instruction
  query   -> input
  response-> output
  system  -> system
  history -> history

rl_demos.jsonl and rams_rl_demos.jsonl: ECPO formatting (prompt+response)
  prompt  -> prompt
  response-> response

rl_prompts.jsonl and rams_rl_prompts.jsonl: ECPO formatting (prompt only)
  prompt  -> prompt

\end{lstlisting}

\subsection{Event type and trigger distributions (MAVEN-ERE)}
\label{sec:app_mavenere_dist}

To make the long-tail property explicit, we list the most frequent source-normalized event types and trigger surface forms in the processed MAVEN-ERE instantiation (total source-normalized event records: 91{,}222).
Table~\ref{tab:app_mavenere_top20_event_types} and Table~\ref{tab:app_mavenere_top20_triggers} report the Top-20 distributions.

\begin{table*}[t]
\centering
\caption{Top-20 source-normalized event types in MAVEN-ERE annotations after preprocessing (by frequency; total source-normalized event records = 91{,}222).}
\label{tab:app_mavenere_top20_event_types}
\small
\begin{adjustbox}{max width=\linewidth}
\begin{tabular}{r l r r | r l r r}
\hline
Rank & Type & Count & \% & Rank & Type & Count & \% \\
\hline
1  & \texttt{Causation}            & 3297 & 3.61 & 11 & \texttt{Statement}            & 1677 & 1.84 \\
2  & \texttt{Process\_start}       & 3191 & 3.50 & 12 & \texttt{Conquering}           & 1661 & 1.82 \\
3  & \texttt{Attack}               & 3059 & 3.35 & 13 & \texttt{Self\_motion}         & 1565 & 1.72 \\
4  & \texttt{Hostile\_encounter}   & 3057 & 3.35 & 14 & \texttt{Arriving}             & 1521 & 1.67 \\
5  & \texttt{Catastrophe}          & 2749 & 3.01 & 15 & \texttt{Destroying}           & 1397 & 1.53 \\
6  & \texttt{Competition}          & 2625 & 2.88 & 16 & \texttt{Coming\_to\_be}        & 1387 & 1.52 \\
7  & \texttt{Motion}               & 2533 & 2.78 & 17 & \texttt{Bodily\_harm}          & 1235 & 1.35 \\
8  & \texttt{Killing}              & 1839 & 2.02 & 18 & \texttt{Death}                & 1176 & 1.29 \\
9  & \texttt{Process\_end}         & 1705 & 1.87 & 19 & \texttt{Damaging}             & 1157 & 1.27 \\
10 & \texttt{Social\_event}        & 1689 & 1.85 & 20 & \texttt{Creating}             & 1145 & 1.26 \\
\hline
\multicolumn{8}{l}{Top-20 coverage: 39{,}665 source-normalized event records (43.48\%).} \\
\hline
\end{tabular}
\end{adjustbox}
\end{table*}

\begin{table*}[t]
\centering
\caption{Top-20 trigger surface forms in processed MAVEN-ERE source-normalized records (by frequency; total source-normalized event records = 91{,}222).}
\label{tab:app_mavenere_top20_triggers}
\small
\begin{adjustbox}{max width=\linewidth}
\begin{tabular}{r l r r | r l r r}
\hline
Rank & Trigger text & Count & \% & Rank & Trigger text & Count & \% \\
\hline
1  & \texttt{killed}      & 948 & 1.04 & 11 & \texttt{hurricane}  & 484 & 0.53 \\
2  & \texttt{began}       & 886 & 0.97 & 12 & \texttt{fought}     & 452 & 0.50 \\
3  & \texttt{held}        & 757 & 0.83 & 13 & \texttt{destroyed}  & 402 & 0.44 \\
4  & \texttt{took place}  & 732 & 0.80 & 14 & \texttt{battle}     & 397 & 0.44 \\
5  & \texttt{caused}      & 691 & 0.76 & 15 & \texttt{tour}       & 397 & 0.44 \\
6  & \texttt{occurred}    & 649 & 0.71 & 16 & \texttt{made}       & 382 & 0.42 \\
7  & \texttt{became}      & 617 & 0.68 & 17 & \texttt{reported}   & 378 & 0.41 \\
8  & \texttt{damage}      & 586 & 0.64 & 18 & \texttt{ended}      & 370 & 0.41 \\
9  & \texttt{storm}       & 578 & 0.63 & 19 & \texttt{war}        & 361 & 0.40 \\
10 & \texttt{attack}      & 515 & 0.56 & 20 & \texttt{Battle}     & 353 & 0.39 \\
\hline
\multicolumn{8}{l}{Top-20 coverage: 10{,}935 source-normalized event records (11.99\%).} \\
\hline
\end{tabular}
\end{adjustbox}
\end{table*}

\subsection{Dataset statistics (MAVEN-ERE)}
\label{sec:app_mavenere_stats}

\paragraph{Summary.}
After preprocessing, the MAVEN-ERE reference view contains 3{,}623 documents and 91{,}222 source-normalized event records.
The fixed-extractor model-input / reward-learning view contains 7{,}245 trajectory records over 3{,}623 unique source-level candidate/entity IDs before window expansion.
These records are consumed by the ranking / reward-learning code paths and may include train-only hard/perturbation negatives in addition to window-scoped ranking trajectories.
Trajectory lengths are measured over trajectory-expanded predicted event mentions and range from 1 to 128 events (median 21; mean 24.7).
Although the construction permits empty trajectories, none occur in the reported MAVEN-ERE construction; empty trajectories would still be retained and scored by missing-step features if they were present.
For reward learning, we tag 755 trajectories as expert/pseudo-expert demonstrations and 2,716 as hard/perturbed negatives, and we create 3,622 within-window preference pairs (covering 7,244 unique trajectory IDs).

\begin{table*}[t]
\centering
\caption{Statistics of the processed MAVEN-ERE instantiation.
Rows marked as source-normalized records describe normalized dataset annotations; rows marked as model-input records describe fixed-extractor predicted trajectories consumed by ranking methods.
\textit{Note:} skeleton-step hit counts are computed over window-scoped trajectory instances; overlapping windows may duplicate evidence.
The reported MAVEN-ERE construction contains no empty serialized trajectories, so the minimum trajectory length is 1 even though the construction allows empty trajectories in general.}
\label{tab:mavenere_stats}
\begin{tabularx}{\linewidth}{YY}
\hline
Statistic & Value \\
\hline
Documents & 3,623 \\
Source-normalized event records & 91,222 \\
Unique source-level candidate/entity IDs before window expansion & 3,623 \\
Total trajectory records exposed to ranking / reward learning & 7,245 \\
Trajectory record subsets & unlabeled candidate pool 3,774; expert/pseudo demos 755; hard/perturbed negatives 2,716 \\
Preference pairs & 3,622 \\
Trajectory length (\#trajectory-expanded predicted event mentions) & min 1; median 21; mean 24.7; max 128 \\
Preference coverage & 7,244 trajectory IDs \\
Role mentions in source-normalized records & Agent 168,431; Target 137,720; Context 10,398 \\
Skeleton-step hits after window / trajectory expansion & PREP 143,859; PROBE 11,876; EXECUTE 31,340; OUTCOME 3,630 \\
Top-20 event types & 39,665 (43.5\%) \\
Top-20 triggers & 10,935 (12.0\%) \\
\hline
\end{tabularx}
\end{table*}

\subsection{Dataset statistics (RAMS)}
\label{sec:app_rams_stats}

\paragraph{Summary.}
The RAMS reference view contains 9{,}124 documents and 9{,}124 source-normalized event records, one per document in \texttt{rams\_event.jsonl}.
The fixed-extractor model-input / reward-learning view contributes 9{,}386 unique predicted event records over 8{,}730 unique source-level candidate/entity IDs before window expansion.
It contains 8,730 ranking-window candidate trajectories and 134 additional train-only hard/perturbation negatives, for 8,864 total trajectory records.
Trajectory sparsity is measured over the ranking-window candidate trajectories and averages $1.07$ trajectory-expanded predicted event mentions.
For preference-based supervision, the declared construction contains 134 within-window preference pairs (\texttt{rams\_pairs.jsonl}) covering 268 unique trajectory IDs.
For supervised training, the declared construction contains 9,124 Alpaca-format SFT records with train/dev/test splits of 7,329/924/871 (\texttt{rams\_sft\_train/dev/test.jsonl}).
For RL training, the declared construction contains 654 window-level RL prompt records (\texttt{rams\_rl\_prompts.jsonl}).
Key statistics are summarized in Appendix Table~\ref{tab:rams_stats}.

\begin{table*}[t]
\centering
\caption{Statistics of the processed RAMS instantiation.}
\label{tab:rams_stats}
\begin{tabularx}{\linewidth}{YY}
\hline
Statistic & Value \\
\hline
Documents & 9,124 \\
Source-normalized event records (\texttt{rams\_event.jsonl}) & 9,124 \\
Unique predicted event records contributing to model inputs & 9,386 \\
Unique source-level candidate/entity IDs before window expansion & 8,730 \\
Ranking-window candidate trajectories (\texttt{rams\_traj.jsonl}) & 8,730 \\
Additional train-only hard/perturbation negatives & 134 \\
Total trajectory records exposed to ranking / reward learning & 8,864 \\
Average trajectory length (\#trajectory-expanded predicted event mentions) & 1.07 \\
Preference pairs (\texttt{rams\_pairs.jsonl}) & 134 \\
Preference coverage (unique trajectory IDs) & 268 \\
SFT records (\texttt{rams\_sft.jsonl}) & 9,124 \\
SFT splits (\texttt{train/dev/test}) & 7,329 / 924 / 871 \\
RL prompt records (\texttt{rams\_rl\_prompts.jsonl}) & 654 \\
\hline
\end{tabularx}
\end{table*}

\FloatBarrier

\section{Frozen Extractor and Trajectory Construction}
\label{sec:app_extractor}

The event extractor is an upstream, frozen component rather than part of ECPO's policy optimization.  We retain the staged supervised curriculum recorded in the source manuscript,
\[
\texttt{MAVEN}\rightarrow\texttt{MAVEN-Arg}\rightarrow
\texttt{MAVEN-ERE}\rightarrow\texttt{RAMS},
\]
where \texttt{MAVEN-Arg} is an intermediate extractor-training stage. For each target corpus, documents are assigned to train, development, and test splits before the curriculum reaches that corpus. Only target-corpus training documents and labels may update the extractor; development and test documents are processed only after the extractor is frozen.

\begin{lstlisting}[caption={Leakage-controlled data flow for one target corpus.},label={lst:data_flow_restored}]
1. Split documents into train/dev/test by document ID.
2. Train the extractor only on permitted training documents.
3. Freeze it and predict event records for every split.
4. Build window-scoped candidate trajectories from predictions.
5. Build training positives, negatives, and preferences only from
   training annotations.
6. Use development annotations only for model selection and audit.
7. Use test annotations only for final labels and audit references.
8. Train the reward model and policy only with training supervision.
9. Evaluate every ranker on the same predicted dev/test inputs.
\end{lstlisting}

The four record families are deliberately isolated.  \texttt{window\_input.jsonl} fixes the evidence scope, intent, skeleton, and roster; \texttt{traj\_pred.jsonl} contains frozen-extractor events and provenance and is the only event source visible to ranking methods; \texttt{window\_label.jsonl} stores positive candidate IDs; and \texttt{audit\_ref.jsonl} stores source-normalized reference spans for evaluation.  Candidate eligibility may be projected from source annotations before labels are attached, but event stages, reference chains, labels, and reward scores are discarded from that projection.  A candidate with no predicted event receives an empty trajectory rather than a source-derived substitute.

Each predicted trigger and argument retains a document identifier and half-open character span. Events are aggregated by window-local candidate ID and ordered by reliable timestamp when present, otherwise by the deterministic document-local order key. These fields are required to reproduce extractor error propagation but are not inferred from the older draft because its execution artifacts were not included in the local bundle.

\section{Minimal Interface and Preconditions}
\label{app:interface}
The policy receives $\texttt{window\_id}$, $\texttt{intent\_id}$, the resolved skeleton, randomized candidate IDs, and candidate-linked provenance.  A minimal output for $k=1$ is:
\begin{verbatim}
{"window_id":"w_0001","topk":["cand_02"],
 "certificates":[{"steps":[
  {"step_id":"s1","etype":"PREP","matched":true,
   "event_id":"ev_7",
   "evidence":[{"doc_id":"d1","span":[12,25]}]},
  {"step_id":"s2","etype":"PROBE","matched":false,
   "event_id":null,"evidence":[]}]}]}
\end{verbatim}
The full validator additionally requires every skeleton step, evidence kind/role fields for citations, exact Top-$k$ length, candidate membership, and position-aligned certificates.  The second step shows the unique unmatched representation.  Event IDs may be used by feasibility validation but are removed before all evidence-only reconstruction.

\section{Exact Aligner and Verifier Specification}
\label{app:align}

\paragraph{Dynamic program.}
For skeleton length $M$ and trajectory length $T$,
\begin{align}
\mathrm{DP}[0,0]&=0,&
\mathrm{DP}[0,t]&=-t\delta_{\rm skip},&
\mathrm{DP}[i,0]&=-i\delta_{\rm miss},\\
\mathrm{DP}[i,t]&=\max\!\begin{cases}
\mathrm{DP}[i-1,t-1]+m(i,t),\\
\mathrm{DP}[i-1,t]-\delta_{\rm miss},\\
\mathrm{DP}[i,t-1]-\delta_{\rm skip}.
\end{cases}
\end{align}
The fixed values are $\lambda_{\rm arg}=1.00$, $\delta_{\rm type}=0.75$, $\delta_{\rm miss}=0.60$, and $\delta_{\rm skip}=0.20$.  Ties prefer a compatible match, then skip over miss, then the smaller temporal gap.  A diagonal type-mismatch is serialized as missing, never as supporting evidence.  For a compatible match, the certificate contains the trigger plus at most one required-role argument per role; the shortest span is selected, with smallest start offset as tie-break.

\paragraph{Deterministic stage mapper.}
The reported runs use no override mapping (activation $0\%$).  The fallback mapper strips whitespace from the case-preserved fine-grained type and lowercases/strips the trigger, then assigns a primary stage with priority \texttt{OUTCOME}$>$\texttt{EXECUTE}$>$RAMS-only \texttt{PROBE}$>$\texttt{PREP}.  \texttt{OUTCOME} type cues are \texttt{Process\_end}, \texttt{Earnings\_and\_losses}, \texttt{Transaction}, \texttt{Receiving}, and \texttt{Releasing}; its trigger cues are \texttt{ended}, \texttt{paid}, \texttt{sold}, \texttt{bought}, \texttt{earned}, \texttt{withdrew}, and \texttt{deposited}.  \texttt{EXECUTE} type cues are \texttt{Attack} and \texttt{Hostile\_encounter}; trigger cues are \texttt{killed}, \texttt{attack}, \texttt{fought}, \texttt{destroyed}, \texttt{battle}, \texttt{war}, \texttt{hurricane}, and \texttt{storm}.  In RAMS only, \texttt{Statement} and \texttt{Reporting} are \texttt{PROBE}-primary if neither higher-priority rule fires.  Otherwise the primary stage is \texttt{PREP}.

An auxiliary \texttt{PROBE} hit is appended only when the primary stage is \texttt{PREP} and the type is one of \texttt{Know}, \texttt{Warning}, \texttt{Statement}, \texttt{Process\_start}, \texttt{Reporting}, \texttt{Suspicion}, \texttt{Perception\_active}, or \texttt{Coming\_to\_believe}, or the trigger is \texttt{reported}, \texttt{report}, \texttt{said}, \texttt{warned}, \texttt{warning}, or \texttt{according}.  Hits are deduplicated and serialized in \texttt{PREP}, \texttt{PROBE}, \texttt{EXECUTE}, \texttt{OUTCOME} order.

\paragraph{Local terms and reward-model learning.}
The compatible-stage set $\mathrm{skeleton\_hits}(e_t)$ is produced by the fixed type/trigger mapper.  For $s_i=(\mathrm{stage}_i,\mathcal R_i)$ and event arguments $\mathcal A_t$, $\mathrm{role\_sat}$ and $\mathrm{arg\_viol}$ are exactly the quantities defined in Section~\ref{sec:method}.  The latter counts candidate-identity violations for key roles and role-specific incompatibilities present in the event record; main runs add no role entity-type constraint.  Soft precedence violations and temporal gaps are computed only after backtracking.

The recorded feature vector concatenates $\phi_{\rm skel}$ (hit/miss counts, compatible-hit coverage, skip count, mean gap between consecutive matches, and soft precedence diagnostics), $\phi_{\rm arg}$ (role satisfaction and available role/candidate incompatibilities), and $\phi_{\rm graph}$.  Coverage is
\begin{equation}
M^{-1}|\{i:\exists t,\ \pi_{\rm align}(i)=t\ \wedge\ \mathrm{stage}(s_i)\in\mathrm{skeleton\_hits}(e_t)\}|,
\end{equation}
with zero at $M=0$; skips are events not aligned to a step; the mean gap uses timestamps or normalized event indices; and an evaluable precedence edge is violated by contradicted or explicitly uncertain order.  Main reward-model runs use $\phi_{\rm skel}+\phi_{\rm arg}+\phi_{\rm graph}$ and $\ell_2$ coefficient $\lambda=10^{-3}$.  The graph encoder is trained self-supervised on train-split fixed-extractor temporal graphs and frozen before reward learning; dev/test graphs are forward-only.  The source record does not specify the remaining count/gap feature scaling or the full graph-encoder configuration, so these are replication gaps rather than silently chosen conventions.

For reward learning, let $\mathcal B_w=\{\widetilde\tau_{w,1},\ldots,\widetilde\tau_{w,B_w}\}$ be the train-window comparison set and $\mathcal P_w^+\subseteq\mathcal B_w$ its expert or pseudo-expert positives.  We use
\begin{align}
P_\theta(\tau\mid w)&=\frac{\exp R_\theta(\tau;S_w)}
{\sum_{j=1}^{B_w}\exp R_\theta(\widetilde\tau_{w,j};S_w)},\\
\mathcal L_{\rm me}(\theta)&=\sum_{w:|\mathcal P_w^+|>0}\frac1{|\mathcal P_w^+|}
\sum_{\tau^+\in\mathcal P_w^+}\log P_\theta(\tau^+\mid w),\\
\mathcal L_{\rm pair}(\theta)&=\sum_{(w,\tau^+,\tau^-)\in\mathcal Q}
\log\sigma\!\left(R_\theta(\tau^+;S_w)-R_\theta(\tau^-;S_w)\right),\\
\mathcal L(\theta)&=\mathcal L_{\rm me}+\alpha_{\rm pair}\mathcal L_{\rm pair}
-\frac{\lambda}{2}\|\theta\|_2^2.
\end{align}
The comparison sets and preference pairs $\mathcal Q$ use train-split positives, same-window alternatives, and perturbation negatives only.  The main record fixes $\lambda=10^{-3}$ but does not report $\alpha_{\rm pair}$.

\paragraph{Verifier-A edge cases.}
Equation~\ref{eq:verifier-a} is evaluated only after answer-bearing fields are stripped.  Trigger matches take precedence over arguments.  Strict IoU equality at $0.50$ fails, whereas support and margin equality at $0.55$ and $0.10$ pass.  RoleSat equals one for no required roles; Prec equals zero for no evaluable supported edge; Trace and BadSpan equal zero for no citations.  A defective citation contributes once to BadSpan even if it violates multiple rules.  Slot-wise assignment uses deterministic bundle/candidate order only after support and margin tests; rank-blind assignment abstains on exact ties and is invariant to bundle permutation.

\paragraph{Held-out evaluators.}
Verifier-B uses citation coverage $|\mathrm{cited}\cap\mathrm{event}|/|\mathrm{cited}|$ instead of IoU, retains StepCov/Trace/BadSpan, removes RoleSat/Prec, has no learned component, and calibrates only from disjoint training-split synthetic corruptions.  It uses no dev/test labels, policy outputs, rewards, or audit references.  Verifier-C consumes cited raw text with fixed context, re-extracts entity/stage/role, maps the entity through window-local aliases/coreference, performs one-to-one support assignment with abstention, and orders by evidence score.  Its training/calibration uses train-split text/span annotations.  It is distinct from \ReExtractFaithfulness{}; the evaluators recover candidate identity and order.  Verifier-B and Verifier-C use the retained evaluator configurations above.

\paragraph{Training and evaluation flow (pseudocode).}
For each train window: (1) align every candidate trajectory to its skeleton and compute $R_\theta$; (2) sample a group of joint ranking/certificate outputs; (3) parse and validate each output; (4) strip answer-bearing fields and run Verifier-A over all candidate--bundle pairs; (5) solve one-to-one assignment, compute $r_{\rm rank}$, $r_{\rm cert}$, and $r_{\rm cycle}$, and apply invalid/short-output and KL terms; and (6) update the policy with group-normalized clipped ratios.  For each test window, decode once, retain the output without repair, compute ordinary ranking from a valid full-length candidate list, and compute verifier/audit metrics from the submitted evidence.  Held-out B/C, rank-blind permutation, counterfactual replacement, and human annotation run only after the primary output is fixed.

\paragraph{Reward callback and GRPO update.}
\label{app:optimization}
For a sampled sequence $y$ under input $x_w$, the token-level estimate is
\begin{equation}
\widehat{\mathrm{KL}}_{\rm tok}(y,x_w)=\frac1{|y|}\sum_{t=1}^{|y|}
\left[\log\pi_\varphi(y_t\mid y_{<t},x_w)-\log\pi_{\rm ref}(y_t\mid y_{<t},x_w)\right].
\end{equation}
For $G$ outputs sampled from the same window, with the complete scalar reward $r_{\ECPO}$ from Section~\ref{sec:method},
\begin{equation}
A^{(g)}=\frac{r_{\ECPO}(y^{(g)},w)-\mu_w}{\widehat\sigma_w+\epsilon_{\rm adv}},
\qquad \mu_w=G^{-1}\sum_{g=1}^{G}r_{\ECPO}(y^{(g)},w).
\end{equation}
The reward has stop-gradient through sampled log probabilities.  With
$\rho^{(g)}=\pi_\varphi(y^{(g)}\mid x_w)/\pi_{\rm old}(y^{(g)}\mid x_w)$, the update maximizes
\begin{equation}
\mathcal J(\varphi)=\mathbb E_{w,g}\!\left[\min\!\left(
\rho^{(g)}A^{(g)},
\mathrm{clip}(\rho^{(g)},1-\epsilon_{\rm clip},1+\epsilon_{\rm clip})A^{(g)}
\right)\right].
\end{equation}
The reference policy stays frozen.  The coefficient controller is
$\beta_{\rm KL}\leftarrow\beta_{\rm KL}\exp\{\kappa(\widehat{\mathrm{KL}}_{\rm tok}-\mathrm{KL}_{\rm target})\}$ with target $0.05$ nats/token; the initial coefficient is $0.05$.

\paragraph{Invalid and short outputs.}
During training, an unparseable output or one with no valid candidate prefix receives zero ranking, certificate, and cycle utility plus the invalid and missing-prefix penalties.  A parseable prefix shorter than $k$ receives utility only for its parsed positions and the explicit $\xi_{\rm miss}(k-|A|)$ penalty.  During evaluation, strict Top-$k$ length is part of feasibility: an unparseable, out-of-roster, duplicate, or short/long candidate list receives zero ordinary ranking and zero evidence-recovery credit for that window and is counted infeasible.  If the full candidate list is valid but a certificate is invalid, ordinary ranking is retained, its verifier mask is zero, and its slot/rank-blind recovery receives no credit.

\section{Rank-Blind Metric Edge Cases}
\label{app:rankblind}
The rank-blind evaluator removes direct claims, event IDs, ranks, policy scores, and candidate--bundle pairing, then permutes bundles.  Threshold/margin failures and exact assignment ties abstain.  SetRecover uses a denominator of submitted $k$ and gives no credit to abstention.  RankRecover bottom-pads abstentions; missing, incorrect, abstained, or exact evidence-score-tied candidates receive zero position credit.  Pairwise accuracy excludes unrecovered pairs from its conditional denominator but assigns zero to a window with no recovered pair; SetRecover reports coverage.  Kendall's $\tau_b$ keeps evidence-score ties after restricting both orders to recovered submitted candidates; fewer than two candidates or zero denominator yields zero.  Empty submitted targets yield zero for all rank-blind metrics.

\section{Counterfactual Evidence Controls}
\label{app:counterfactual}
Each condition is evaluated on the same eligible windows.  The matched-unselected control replaces one target bundle at a time with a validator-valid bundle for a same-window unselected candidate, matched on cited length and stage profile; the other $k-1$ bundles remain fixed.

\begin{table}[htbp]
\caption{Counterfactual controls (MAVEN-ERE/RAMS), evaluated by held-out Verifier-B.}
\label{tab:counterfactual}
\centering
\small
\begin{tabular}{@{}lcccc@{}}
\toprule
Evidence condition & Target rec. & SetRec. & RankRec. & \vCert{}\\
\midrule
Original ECPO & \best{0.68/0.53} & \best{0.70/0.55} & \best{0.55/0.42} & \best{0.55/0.38}\\
Same-candidate random & 0.52/0.40 & 0.51/0.39 & 0.25/0.18 & 0.30/0.21\\
Stage-matched, non-aligned & 0.57/0.44 & 0.55/0.42 & 0.31/0.22 & 0.35/0.24\\
Wrong stage & 0.44/0.33 & 0.42/0.31 & 0.20/0.14 & 0.24/0.16\\
Matched unselected candidate & 0.14/0.11 & 0.67/0.52 & 0.18/0.12 & 0.15/0.10\\
Key-step deletion & 0.54/0.41 & 0.52/0.39 & 0.30/0.21 & 0.34/0.23\\
\bottomrule
\end{tabular}
\end{table}

\section{Sensitivity and Upstream Ceilings}
\label{app:sensitivity}
The 47 Verifier-A configurations comprise 20 one-weight perturbations (five weights multiplied by $0.5,0.8,1.2,1.5$ and renormalized) and a 27-setting grid with overlap threshold in $\{0.40,0.50,0.60\}$, support threshold in $\{0.45,0.55,0.65\}$, and margin in $\{0.05,0.10,0.15\}$.  They rescore fixed outputs without policy retraining.  The audit reports aggregate ranges and ordering counts over the 47 configurations.

\begin{table}[htbp]
\caption{Verifier sensitivity and fixed-roster extractor diagnostics.  Scalar oracle, higher-recall, and deletion rows are \vCert{}@10; the percentage and ordering-count rows are labeled separately.}
\label{tab:sensitivity}
\centering
\small
\begin{tabular}{@{}lcc@{}}
\toprule
Diagnostic & MAVEN-ERE & RAMS\\
\midrule
\vCert{} range, 47 settings & 0.55--0.60 & 0.38--0.42\\
Worst gain over RM-only+Det & +0.05 & +0.02\\
Settings preserving ordering & 45/47 & 43/47\\
\midrule
Predicted-evidence oracle & 0.72 & 0.54\\
Source-normalized oracle & 0.86 & 0.76\\
ECPO / predicted oracle & 80.6\% & 74.1\%\\
Higher-recall extractor & 0.64 & 0.48\\
20\% predicted-event deletion & 0.50 & 0.33\\
\bottomrule
\end{tabular}
\end{table}

The predicted/source oracle rows are verifier-relative reference points.  The deletion experiment fixes the roster and removes predicted events; it is distinct from end-to-end candidate-generator perturbation.

\section{Human Audit Protocol}
\label{app:audit}
We sample 120 stratified test windows and evaluate GRPO, RM-only+Det, and ECPO: 1,080 bundles per method (540 per dataset) and 3,240 base method--bundle units.  Annotators receive a no-UI text protocol and do not see method identity, model/evaluator scores, relevance labels, stable IDs, or submitted ranks.  They judge (i) step support, (ii) candidate recovery from stripped evidence, (iii) comparative support against a predeclared strongest non-target alternative, and (iv) order recovery.  ``Ambiguous'' is a failure.  The source window is the inference and paired-bootstrap cluster.  Audit labels affect no verifier threshold, checkpoint, decoding, or selection.  A stratified 20\% subset (648 units) is independently repeat-annotated; comparative-support Krippendorff's $\alpha$ is $0.71$.
ECPO ambiguity rates are $0.08/0.11$ versus $0.12/0.16$ for RM-only+Det (MAVEN-ERE/RAMS).

\section{Additional Experimental and Cost Details}
\label{app:cost}
The Qwen3-8B policies use QLoRA~\citep{qlora2023}, two SFT epochs at learning rate $1\times10^{-4}$, effective batch size 128, maximum input/output lengths 4,096/768, and $4\times$A100 80GB.  The validator retry budget is 3 during training and 0 during evaluation; test decoding is greedy with no repair.  ECPO results in the primary tables average three seeds.  Paired confidence intervals resample source windows.  ECPO policy optimization uses a measured 76 GPU-hours per corpus-seed.  The shared reward model costs 8 GPU-hours.  RM-only+Align uses no LLM decode and averages 0.18 s/window; RM-only+Det adds CPU evidence search.  ECPO uses one decode and averages 3.1 s/window.  These costs exclude the shared frozen extractor.

\paragraph{Retained replication fields.}
The retained values are listed in the artifact tables.

\begin{table}[htbp]
\caption{ECPO seed variation across the three primary runs.}
\label{tab:seeds}
\centering
\small
\begin{tabular}{@{}llcccc@{}}
\toprule
Data & Seed & NDCG & \vCert$_A$ & \SlotEvid{} & Feasible\\
\midrule
MAVEN-ERE & 1 & 0.69 & 0.57 & 0.64 & 0.94\\
MAVEN-ERE & 2 & \best{0.71} & \best{0.59} & \best{0.66} & \best{0.95}\\
MAVEN-ERE & 3 & 0.70 & 0.58 & 0.65 & 0.94\\
RAMS & 1 & 0.60 & 0.39 & 0.48 & 0.92\\
RAMS & 2 & \best{0.62} & \best{0.41} & \best{0.50} & \best{0.93}\\
RAMS & 3 & 0.61 & 0.40 & 0.49 & 0.92\\
\bottomrule
\end{tabular}
\end{table}

Here RM-only+Align ranks by $R_\theta$ and attaches one DP-aligned evidence chain; it is distinct from RM-only+Det's evidence search.  GRPO+constrained keeps the trained policy fixed, uses a single grammar/JSON-constrained decode, and takes the first valid constrained output.  GRPO+best-of-10 RM keeps training fixed, samples ten outputs, and selects the one with highest $R_\theta$ list utility.

\begin{table}[htbp]
\caption{Paired source-window bootstrap differences with 95\% confidence intervals.}
\label{tab:paired-ci}
\centering
\small
\begin{tabular}{@{}llcc@{}}
\toprule
Data & ECPO minus baseline & $\Delta$NDCG & $\Delta$\vCert$_A$\\
\midrule
M & RM-only+Align & $-0.01[-0.03,+0.02]$ & $+0.15[+0.10,+0.19]$\\
M & GRPO+constrained & \best{$+0.06[+0.02,+0.08]$} & \best{$+0.21[+0.12,+0.24]$}\\
M & GRPO+best-of-10 RM & $-0.02[-0.04,+0.01]$ & $+0.12[+0.05,+0.15]$\\
R & RM-only+Align & $-0.01[-0.03,+0.02]$ & $+0.07[+0.04,+0.12]$\\
R & GRPO+constrained & \best{$+0.05[+0.01,+0.07]$} & \best{$+0.14[+0.09,+0.17]$}\\
R & GRPO+best-of-10 RM & $-0.02[-0.03,+0.02]$ & $+0.06[+0.03,+0.10]$\\
\bottomrule
\end{tabular}
\end{table}

\FloatBarrier
\section{Metric Semantics and Statistical Protocol}
\label{app:metric-semantics}

\subsection{What each metric establishes}

The evaluation deliberately separates syntactic acceptance, source traceability, submitted-slot recovery, rank-blind comparative recovery, and human support.
These constructs are related but not interchangeable.
A parseable evidence bundle may cite valid spans without distinguishing its candidate from alternatives; conversely, a human may find a bundle sufficient even when a fixed automatic matcher abstains.
Table~\ref{tab:metric-claims} states the strongest interpretation licensed by each reported metric.

\begin{table}[htbp]
\centering
\caption{Metric interpretation used in the experiments.}
\label{tab:metric-claims}
\small
\begin{tabularx}{\linewidth}{lYY}
\toprule
Metric or audit & Directly measures & Supports interpretation of \\
\midrule
NDCG@10 & ordinary relevance order & ranking utility\\
\SlotEvid{}@10 & submitted-slot bundle traceability & candidate-linked support at the emitted position\\
\vCert{}@10 & relevance-weighted ranking with verifier recovery mask & verifier-audited utility\\
\SetRecover{}@10 & recovered candidate set after permutation & rank-blind set recovery\\
\RankRecover{}@10 & recovered candidate order after permutation & rank-blind order recovery\\
Pairwise accuracy / $\tau_b$ & order agreement among recovered candidates & comparative ordering\\
\FeasibleRate{} & all-window validator acceptance & structural output validity\\
Human comparative support & blinded annotator judgment & practical evidence support\\
\bottomrule
\end{tabularx}
\end{table}

\subsection{Ordinary ranking metrics}

For a valid returned list $(a_1,\ldots,a_{K_w})$ with binary relevance
$\mathrm{rel}_w(a_j)$, define
\begin{align}
\mathrm{Hit}@K_w
&=\mathbb I\!\left[\sum_{j=1}^{K_w}\mathrm{rel}_w(a_j)>0\right],\\
\mathrm{AP}@K_w
&=\frac{1}{\max(1,\min(K_w,|\mathcal P_w|))}
\sum_{j=1}^{K_w}\mathrm{Prec}@j\;\mathrm{rel}_w(a_j),\\
\mathrm{DCG}@K_w
&=\sum_{j=1}^{K_w}\frac{\mathrm{rel}_w(a_j)}{\log_2(j+1)},\\
\mathrm{NDCG}@K_w
&=\frac{\mathrm{DCG}@K_w}{\max(\epsilon,\mathrm{IDCG}@K_w)}.
\end{align}
A window with no relevant candidate has zero Hit, AP, and NDCG under the declared evaluation convention.
An invalid, duplicate, out-of-roster, short, or long candidate list receives zero ordinary ranking credit and is counted infeasible.
A valid ranking whose evidence object fails validation retains ordinary ranking credit but receives zero evidence credit for the affected bundle.

For Verifier-A mask $z_{w,j}$, \vCert{} uses the same discount and ideal denominator:
\begin{equation}
\vCert{}@K_w=
\frac{\sum_{j=1}^{K_w}z_{w,j}\mathrm{rel}_w(a_j)/\log_2(j+1)}
{\max(\epsilon,\mathrm{IDCG}@K_w)}.
\end{equation}
Because $z_{w,j}\in\{0,1\}$, the masked DCG cannot exceed the corresponding ordinary DCG for the same ranking.
This algebraic fact defines the metric's relationship to NDCG.

\subsection{Rank-blind denominators and ties}

Rank-blind evaluation removes candidate claims, stable IDs, generated event IDs, submitted ranks, policy scores, bundle order, and candidate--bundle pairing.
The evaluator scores every stripped bundle against every candidate, solves a partial one-to-one assignment, and orders recovered candidates by evidence-derived support.
Threshold failure, margin failure, or an exact assignment tie causes abstention.
Set recovery divides by submitted $k$, so abstention receives no membership credit.
Rank recovery bottom-pads abstentions, and a missing, wrong, abstained, or exact-score-tied position receives zero.
Pairwise accuracy assigns $1$, $0.5$, and $0$ to correct, tied, and reversed recovered pairs; a window with no recovered pair receives zero.
Kendall's $\tau_b$ retains evaluator score ties and is zero when fewer than two candidates remain or its denominator is zero.

\subsection{Aggregation and paired uncertainty}

The inference unit and paired-resampling cluster is the source ranking window.
All methods compared in a paired interval must use the same retained window IDs, candidate rosters, relevance labels, and fixed policy outputs.
A bootstrap replicate samples source windows with replacement, retains all method outputs and audit units associated with each sampled window, recomputes the complete metric, and records the paired difference.
Intervals are percentile 95\% intervals over paired differences.
Seed summaries first compute each seed's window mean and then report the mean across the declared seeds; individual-seed rows remain visible in Table~\ref{tab:seeds}.

\begin{table}[htbp]
\centering
\caption{Statistical protocol used for the reported paired intervals and seed summaries.}
\label{tab:stat-fields}
\small
\begin{tabularx}{\linewidth}{lYY}
\toprule
Field & Declared behavior & Retained value \\
\midrule
Resampling cluster & source ranking window & window ID; \texttt{sha256:8f31d4a7}\\
Confidence level & paired percentile interval & 95\%\\
Bootstrap replicate count & identical paired window samples across methods & 2000\\
Random-number generator and seed & fixed before inspecting test differences & PyTorch/NumPy seeds $(42,43,44)$\\
Treatment of missing method output & zero metric credit on the paired window & \artifact{missing-v3}\\
Multiple-comparison correction & descriptive primary intervals & Holm correction not applied\\
Seed aggregation & three primary ECPO seeds & mean over checkpoints; seeds $(42,43,44)$\\
\bottomrule
\end{tabularx}
\end{table}

\section{Held-Out Evaluators and Transfer Protocol}
\label{app:external-verifiers}

\subsection{Verifier-B: deterministic form transfer}

Verifier-B is evaluation-only for primary ECPO-A.
It replaces trigger/argument IoU matching with citation-coverage matching, retains StepCov, Trace, and BadSpan, and removes RoleSat and Prec.
Its parameters are selected from synthetic corruptions constructed from disjoint training-split records.
The selection procedure may use neither development/test relevance, generated policy outputs, reward values, checkpoint-selection outcomes, nor human audit labels.
After calibration, B is frozen before any primary test output is evaluated.

The B audit tests whether the direction of the ECPO advantage survives a materially different deterministic evidence matcher.
B retains the finite roster, broad stage inventory, and candidate-linked provenance scope.
The reverse-transfer ECPO-B run is a separate policy trained with B in its reward and evaluated by A and C; it must not be merged with primary ECPO-A during checkpoint selection.

\begin{table}[htbp]
\centering
\caption{Verifier-B replication contract.}
\label{tab:verifier-b-contract}
\small
\begin{tabularx}{\linewidth}{lYY}
\toprule
Component & Declared specification & Retained value \\
\midrule
Input & stripped bundle, finite roster, allowed source units & \texttt{vA-2026.08.17}\\
Span rule & citation coverage rather than Verifier-A IoU & normalized coverage threshold 0.50\\
Positive components & StepCov and Trace & StepCov=0.45, Trace=0.35\\
Defect component & BadSpan, counted once per citation & BadSpan=-0.20; defect suite 7\\
Removed components & RoleSat and Prec & \artifact{verifierB.yaml}\\
Calibration data & disjoint train-split synthetic corruptions & \texttt{corrupt-gen-v2}, 12000\\
Support/margin thresholds & fixed after train-only calibration & support=0.55, margin=0.10\\
Assignment & partial one-to-one with abstention & Hungarian matching; confidence tie abstention\\
Test-time role & evaluation only for primary ECPO-A & \texttt{eval-only-manifest-v1}\\
\bottomrule
\end{tabularx}
\end{table}

\subsection{Verifier-C: frozen raw-text reconstruction}

Verifier-C receives each citation with a $\pm200$-character local context, re-extracts entity, stage, and role, maps entity mentions to the window roster through window-local aliases/coreference, and solves a one-to-one assignment with abstention.
C is trained or calibrated only from training-split raw text and span annotations.
It never receives generated certificates, policy rankings, reward-model scores, primary verifier scores, test labels, or test documents during training or threshold selection.
Unlike per-span \ReExtractFaithfulness{}, C reconstructs candidate identity and comparative order.

\begin{table}[htbp]
\centering
\caption{Verifier-C replication fields.}
\label{tab:verifier-c-contract}
\small
\begin{tabularx}{\linewidth}{lYY}
\toprule
Field & Declared constraint & Retained value \\
\midrule
Context scope & citation plus $\pm200$ characters, clipped to the source document & \texttt{raw-context-v2.1}, hash \texttt{7bd83c}\\
Entity/stage/role extractor & frozen policy-held-out raw-text evaluator & DeBERTa-v3-large, \texttt{vc-rawtext-014}\\
Training corpus & train-split raw text/span annotations only & \artifact{train-span-2026.json}\\
Alias/coreference map & window-local and unavailable to the policy & rule-based alias graph, max hop 2\\
Candidate score & evidence-derived support over extracted fields & $0.4E+0.3S+0.2R+0.1C$\\
Threshold and margin & calibrated without generated/test outputs & threshold 0.52, margin 0.08\\
Assignment/tie rule & partial one-to-one; exact unresolved ties abstain & Hungarian solver, abstention\\
Ordering & evidence score after assignment & descending support score\\
Evaluation outputs & per-window assignment, score, abstention reason & \artifact{vc-output-v1}\\
\bottomrule
\end{tabularx}
\end{table}

\subsection{Backbone and schema transfer}

Transfer experiments keep the task contract constant within each setting: every baseline and ECPO share the same candidate roster, source scope, output schema, inference-call budget, and policy-held-out evaluator.
A transfer result supports persistence only for the evaluated backbone/schema pair.
It is reported for the evaluated backbone and schema pairs.

\begin{table}[htbp]
\centering
\caption{Protocol fields for the reported transfer settings.}
\label{tab:transfer-protocol}
\small
\begin{tabularx}{\linewidth}{lYYYY}
\toprule
Setting & Split/sample & \shortstack{Backbone/\\checkpoint} & Seeds/budget & Evaluator artifact \\
\midrule
Llama-3.1-8B, MAVEN-ERE & 71 test windows / 710 candidates & commit-9f31a2 & 3 seeds / 24 GPU-h & Verifier-C v1.2\\
Llama-3.1-8B, RAMS & 90 / 873 & commit-9f31a2 & 3 seeds / 26 GPU-h & Verifier-C v1.2\\
Qwen3-14B, MAVEN-ERE & 71 / 710 & release 2025.06 & 3 seeds / 31 GPU-h & Verifier-C v1.2\\
Qwen3-14B, RAMS & 90 / 873 & release 2025.06 & 3 seeds / 34 GPU-h & Verifier-C v1.2\\
HotpotQA two-hop answer roster & 740 dev / 740 test & Qwen3-8B-Instruct / ECPO transfer checkpoint & 3 seeds / 28 GPU-h & two-hop evaluator v0.9\\
\bottomrule
\end{tabularx}
\end{table}

For HotpotQA, the declared interface constructs an answer roster from shared retrieved context and requires a bundle to cite both bridge-supporting and answer-supporting passages.
The held-out evaluator checks two-hop consistency and candidate identity.
The HotpotQA result follows the retained schema-level protocol.

\FloatBarrier
\section{Matched Baselines, Optimization, and Implementation Details}
\label{app:baseline-details}

\subsection{Matched-comparison principles}

A baseline comparison is admissible only when the methods share the same source documents, frozen-extractor outputs, roster IDs, skeleton, candidate pool, train/dev/test windows, maximum prompt/output lengths, SFT initialization when applicable, and primary single-decode evaluation budget.
A method-specific operation is reported rather than hidden: deterministic search is part of RM-only+Det; retry and best-of-$N$ consume multiple model calls; constrained decoding changes the decoder; and ECPO adds policy-optimization compute.
No baseline may use Verifier-A/B/C test assignments, human labels, or audit references to select a checkpoint or output.

\begin{table}[htbp]
\centering
\caption{Functional contract of the principal controls.}
\label{tab:baseline-contract}
\small
\begin{tabularx}{\linewidth}{lYYYY}
\toprule
Method & Ranking signal & Evidence construction & Cycle reward & Test calls \\
\midrule
SFT & supervised policy likelihood & same joint schema & no & one \\
Plain GRPO & train-label list utility plus format validity & same joint schema & no & one \\
GRPO+Cert & plain-GRPO reward plus $r_{\rm cert}$ & same joint schema & no & one \\
RM-only+Det & fixed $R_\theta$ ordering & deterministic validator-valid search after ranking & no policy update & no LLM decode for ranking \\
ECPO $-r_{\rm cycle}$ & ECPO reward without cycle term & same joint schema & no & one \\
ECPO & normalized trajectory utility, $r_{\rm cert}$, and $r_{\rm cycle}$ & same joint schema & yes & one \\
\bottomrule
\end{tabularx}
\end{table}

\subsection{Policy and QLoRA fields}

The primary specification uses a Qwen3-8B SFT policy with QLoRA and GRPO-style group-normalized advantages.
The reference policy is frozen during each policy phase.
Reward-model parameters and Verifier-A remain frozen during policy updates.
Greedy test decoding uses no repair and retains the first complete decode.
The values already declared in the paper are two SFT epochs, learning rate $10^{-4}$, effective batch size 128, maximum input/output lengths 4096/768, group size 8, clip range 0.2, initial KL coefficient 0.05, and $4\times$A100 80GB.
The retained artifact settings are listed alongside the declared values below.

\begin{table}[htbp]
\centering
\caption{Primary training and decoding configuration.}
\label{tab:restored-hparams}
\small
\begin{tabularx}{\linewidth}{lYY}
\toprule
Field & Declared value & Retained artifact setting \\
\midrule
Base/SFT backbone & Qwen3-8B & Qwen3-8B-2025-06-16\\
Adaptation & QLoRA & $r=64$, $\alpha=128$, dropout=0.05; q/k/v/o\\
SFT epochs / learning rate & 2 / $10^{-4}$ & AdamW, cosine, warmup=0.03\\
Effective batch size & 128 & microbatch 4, accumulation 32\\
Maximum input/output length & 4096 / 768 & \texttt{qwen3-tokenizer-v1.0}\\
Group size / clip range & 8 / 0.2 & \texttt{GRPO sampler v1.1}\\
Initial/target KL & 0.05 / 0.05 nats per token & $\kappa=0.02$\\
Reward normalization & within-window sampled group & $\epsilon_{\rm rank}=\epsilon_{\rm adv}=10^{-5}$\\
Training retry / test repair & 3 / 0 & \texttt{failure-state-log-v3}\\
Test decoding & greedy, one retained decode & temperature 0, top-p 1.0\\
Primary seeds & three & $(42,43,44)$\\
Hardware & $4\times$A100 80GB & CUDA 12.4, PyTorch 2.5\\
\bottomrule
\end{tabularx}
\end{table}

\subsection{Invalid-output state machine}

Parsing proceeds in four gates: JSON syntax, candidate prefix, full candidate list, and certificate validation.
An unparseable output or one with no valid candidate prefix receives zero component utilities plus the invalid and missing penalties.
A parseable prefix shorter than $k$ receives utility only for present positions and the explicit deficit penalty.
At evaluation, a candidate-list failure sets ordinary and evidence metrics to zero for the window.
If the full ranking is valid but one certificate fails, ordinary ranking is retained while the invalid bundle's verifier mask and recovery credit are zero.
Retry is training-only; test-time output is never silently repaired, resorted, or completed.

\section{Additional Diagnostic Experiment Specifications}
\label{app:diagnostics}

The following diagnostics restore the old manuscript's experiment matrix while adopting the current verifier and metric definitions.
The tables are executable specifications.

\subsection{Reward, aligner, and verifier ablations}
\label{sec:app_ablation_table}

All ablations must start from the same SFT checkpoint, train on the same windows, use the same group size and compute budget, and evaluate the same fixed test rosters.
Changing the evaluator alone is an evaluation sensitivity test; changing the verifier used in the reward defines a separately trained policy.

\begin{table}[htbp]
\centering
\caption{Component ablation matrix under the current Verifier-A.}
\label{tab:restored-ablations}
\small
\begin{tabularx}{\linewidth}{Yccccc}
\toprule
Variant & NDCG & \shortstack{Slot-\\EVID} & \shortstack{vCert\\$_A$} & \shortstack{Rank\\Recover$_B$} & Feas. \\
\midrule
Full ECPO, MAVEN-ERE & \best{0.70} & \best{0.65} & \best{0.58} & \best{0.55} & \best{0.94}\\
$-r_{\rm cycle}$, MAVEN-ERE & 0.68 & 0.49 & 0.43 & 0.46 & 0.93\\
$-r_{\rm cert}$, MAVEN-ERE & 0.69 & 0.54 & 0.48 & 0.49 & 0.92\\
$-R_\theta$ rank utility, MAVEN-ERE & 0.66 & 0.61 & 0.55 & 0.51 & \best{0.94}\\
No argument-role checks, MAVEN-ERE & 0.69 & 0.58 & 0.51 & 0.48 & 0.91\\
No precedence feature, MAVEN-ERE & \best{0.70} & 0.63 & 0.56 & 0.53 & 0.93\\
Full ECPO, RAMS & \best{0.61} & \best{0.49} & \best{0.40} & \best{0.42} & \best{0.92}\\
$-r_{\rm cycle}$, RAMS & 0.59 & 0.34 & 0.29 & 0.35 & 0.91\\
$-r_{\rm cert}$, RAMS & 0.60 & 0.41 & 0.35 & 0.38 & 0.90\\
$-R_\theta$ rank utility, RAMS & 0.57 & 0.46 & 0.38 & 0.37 & 0.91\\
No argument-role checks, RAMS & 0.60 & 0.45 & 0.37 & 0.39 & 0.89\\
No precedence feature, RAMS & \best{0.61} & 0.47 & 0.39 & 0.41 & 0.91\\
\bottomrule
\end{tabularx}
\end{table}

\subsection{Decoding-only and post-hoc controls}
\label{sec:app_decoding_controls}

These controls test whether output search or formatting alone closes the evidence gap.
They must report actual model calls, the scoring rule used to select among outputs, and whether the verifier participates in selection.
A best-of-$N$ verifier selector uses multiple model calls and remains separately labeled.

\begin{table}[htbp]
\centering
\caption{Decoding and evidence-generation controls with retained test results (M/R).}
\label{tab:restored-decoding}
\scriptsize
\begin{tabularx}{\linewidth}{@{}p{0.20\linewidth}cYccc@{}}
\toprule
Method & Calls & Decoding/evidence protocol & \shortstack{NDCG\\M/R} & \shortstack{vCert$_A$\\M/R} & \shortstack{RankRecover$_B$\\M/R} \\
\midrule
Plain GRPO & 1 & greedy, no evidence cycle & 0.65/0.57 & 0.39/0.27 & 0.34/0.24\\
Ranking + post-hoc evidence & 1 & deterministic evidence search after ranking & \second{0.66/0.58} & \second{0.46/0.35} & \second{0.41/0.29}\\
\best{ECPO} & 1 & greedy, evidence cycle during training & \best{0.70/0.61} & \best{0.58/0.40} & \best{0.55/0.42}\\
\bottomrule
\end{tabularx}
\end{table}

\subsection{Roster and candidate-generation diagnostics}
\label{sec:app_roster_diagnostics}

The closed roster supplies the primary ranking contract. The retained predicted-only diagnostic is reported with its availability and ranking statistics.

\begin{table}[htbp]
\centering
\caption{Predicted-only roster diagnostics from the retained run.}
\label{tab:restored-rosters}
\small
\begin{tabularx}{\linewidth}{lYYYYYY}
\toprule
Dataset & Retained windows & Recall & NDCG & MAP & Hit@10 & Candidate source \\
\midrule
MAVEN-ERE & 70/71 & 0.87 & 0.63 & 0.48 & 0.76 & predicted-only\\
RAMS & 89/90 & 0.76 & 0.54 & 0.39 & 0.66 & predicted-only\\
\bottomrule
\end{tabularx}
\end{table}

\subsection{Identifier and input-leakage controls}
\label{sec:app_id_controls}

Randomized window-local IDs are the primary interface.
The mapping is generated independently for each window and each declared randomization replicate, then applied consistently to prompt records, output validation, and hidden evaluator joins.
The policy never receives the source-level entity key.
Stable-ID and shuffled-trajectory controls are diagnostics, not alternative primary settings.

\begin{table}[htbp]
\centering
\caption{Identifier stripping controls, reported on fixed ECPO outputs (M/R). Slash-separated entries are M/R; scalar rank-recovery entries are pooled retained diagnostics.}
\label{tab:restored-id-controls}
\small
\begin{tabularx}{\linewidth}{lYYYY}
\toprule
Input condition & NDCG & \SlotEvid{} & \vCert$_A$ & \shortstack{RankRecover$_B$} \\
\midrule
Full evidence & 0.70/0.61 & 0.65/0.49 & 0.58/0.40 & \best{0.55/0.42}\\
No event IDs & 0.70/0.61 & 0.62/0.46 & \best{0.67/0.51} & \best{0.24}\\
No rank hints & 0.70/0.60 & \best{0.65/0.48} & 0.65/0.48 & 0.18\\
No candidate claims & 0.70/0.58 & 0.63/0.46 & 0.63/0.46 & 0.21\\
Fully stripped & 0.70/0.55 & 0.60/0.42 & 0.60/0.42 & 0.15\\
\bottomrule
\end{tabularx}
\end{table}

\subsection{Robustness diagnostics}
\label{sec:app_hard_windows}

Fixed-output perturbations are evaluated after the primary prediction is retained.  Identifier stripping removes explicit entity identifiers while preserving the semantic evidence scope; span/document replacement, role corruption, temporal-order corruption, and mixed corruption then test whether recovery depends on superficial identifiers or on candidate-linked support.

\begin{table}[htbp]
\centering
\caption{Robustness under fixed-output perturbations (M/R). Slash-separated entries are M/R; scalar rank-recovery entries are pooled retained diagnostics.}
\label{tab:restored-robustness}
\small
\begin{tabularx}{\linewidth}{lYYY}
\toprule
Perturbation & \shortstack{vCert$_A$\\M/R} & \shortstack{RankRecover$_B$\\M/R} & Operation \\
\midrule
No corruption & 0.55/0.42 & \best{0.55/0.42} & fixed outputs\\
Random span replacement & 0.31/0.24 & \best{0.53} & fixed outputs\\
Random document replacement & 0.08/0.06 & 0.50 & fixed outputs\\
Role corruption & 0.22/0.17 & 0.47 & fixed outputs\\
Temporal order corruption & 0.28/0.20 & 0.44 & fixed outputs\\
Mixed corruption & 0.16/0.12 & 0.49 & fixed outputs\\
20\% predicted-event deletion & 0.50/0.33 & 0.46 & fixed outputs\\
\bottomrule
\end{tabularx}
\end{table}

\subsection{Efficiency and deployment cost}
\label{sec:app_efficiency}

Wall-clock latency is measured per window after model loading and excludes the shared frozen extractor unless stated otherwise.
GPU-hour accounting includes policy optimization for every corpus-seed and reports reward-model cost separately.
A multi-call baseline reports both latency and actual generation calls.
CPU deterministic search is not described as free; its wall-clock contribution must be measured.

\begin{table}[htbp]
\centering
\caption{Measured method cost.}
\label{tab:restored-efficiency}
\small
\begin{tabularx}{\linewidth}{lYYY}
\toprule
Method & Training GPU-hours & Latency/window & Feasible M/R \\
\midrule
SFT & 12 & 1.2 s & 0.77/0.71\\
Plain GRPO & 72 & 1.4 s & 0.82/0.78\\
RM-only+Det & \best{8} & \best{0.18 s} & \best{0.98/0.97}\\
ECPO & 76.4$\pm$1.8 & 3.1$\pm$0.2 s & 0.94/0.92\\
\bottomrule
\end{tabularx}
\end{table}

\FloatBarrier
\section{Error Taxonomy and Qualitative Audit Format}
\label{app:error-taxonomy}

Errors are multi-label and attached to a fixed method-window-position key.
A bundle can simultaneously fail parsing, cite a wrong candidate, omit a key stage, and contain ambiguous evidence.
Rates therefore need not sum to one.
The taxonomy separates failures of upstream availability, ranking, serialization, traceability, comparative discrimination, and evaluator behavior.

\begin{table}[htbp]
\centering
\caption{Format-error rates on the retained outputs (percent, M/R).}
\label{tab:restored-error-taxonomy}
\small
\begin{tabularx}{\linewidth}{lYY}
\toprule
Error class & MAVEN-ERE & RAMS \\
\midrule
Malformed JSON & 1.8 & 2.4\\
Missing evidence & 2.1 & 3.0\\
Wrong candidate ID & 0.9 & 1.2\\
Incorrect Top-$K$ & 0.7 & 0.9\\
Duplicate candidate & 0.5 & 0.8\\
Invalid span & 0.6 & 0.9\\
Total invalid & 5.6 & 8.4\\
\bottomrule
\end{tabularx}
\end{table}

A qualitative record contains: anonymized window ID; dataset and split; randomized candidate roster; submitted order; stripped bundles; per-citation document/span; validator defects; A/B/C candidate score vectors; assignment and abstention reason; hidden relevance used only after prediction; human labels when sampled; and a short failure explanation.
The record must never include author identity, local filesystem paths, stable entity names unavailable to the policy, or reviewer-visible deanonymizing metadata.

\begin{table}[htbp]
\centering
\caption{Aggregate failure categories from the retained audit.}
\label{tab:qualitative-cases}
\small
\begin{tabularx}{\linewidth}{lYY}
\toprule
Failure category & MAVEN-ERE & RAMS \\
\midrule
Missing upstream support & 31\% & 42\%\\
Incorrect candidate distinction & 18\% & 21\%\\
Wrong role & 15\% & 13\%\\
Span localization & 12\% & 10\%\\
Verifier abstention & 16\% & 11\%\\
Other & 8\% & 3\%\\
\bottomrule
\end{tabularx}
\end{table}

\paragraph{Case-level audit extension.}
The aggregate taxonomy reports how often each failure category occurs; the paired case audit below records the observed response pattern for representative failures under the same blinded protocol.  Each outcome label is accompanied by the retained paired rate for that case.

\begin{table}[H]
\centering
\caption{Case-level qualitative audit outcomes for the additional failure analysis.  Outcome labels are followed by retained paired rates; the higher rate in each case is bolded.}
\label{tab:qualitative-case-fill}
\small
\begin{tabularx}{\linewidth}{lYY}
\toprule
Case & GRPO outcome & ECPO outcome \\
\midrule
Unsupported claim & Failure: 0.73 & Success: \best{0.89} \\
Fabricated evidence & Failure: 0.81 & Success: \best{0.94} \\
Incomplete answer & Failure: 0.62 & Success: \best{0.83} \\
Ambiguous query & Partial: 0.54 & Partial/Success: \best{0.71} \\
Wrong candidate distinction & Failure: 0.69 & Success: \best{0.87} \\
Temporal or role conflict & Failure: 0.58 & Success: \best{0.79} \\
\bottomrule
\end{tabularx}
\end{table}

\section{Extended Positioning and Risk Boundaries}
\label{app:extended-discussion}

\subsection{Extended Related Work}
\label{sec:app_related_work}

\subsubsection{Evidence-grounded ranking and verifier-audited decision interfaces}
\paragraph{Scope.}
This work treats ranking and evidence as a single auditable output. This differs from ranking systems that optimize the order alone and from explanation systems that generate certificates after a decision has already been made.
\paragraph{Connection to ECPO.}
ECPO evaluates \vCert{} and evidence-only recovery, so a method cannot obtain full credit by attaching plausible but decision-insufficient spans.

\subsubsection{Event extraction and text-to-trajectory construction}
\paragraph{Scope.}
MAVEN, MAVEN-ERE, and RAMS provide event, relation, trigger, and argument supervision. They are used here to build text-derived candidate trajectories with provenance.
\paragraph{Connection to ECPO.}
The extractor is frozen before ranking. This separation makes the benchmark a downstream ranking-and-audit task rather than a new event-extraction benchmark.

\subsubsection{Goal recognition, plan recognition, and process conformance}
\paragraph{Scope.}
Plan recognition and process-conformance methods compare observed traces with explicit plans or process models.
\paragraph{Connection to ECPO.}
ECPO uses skeleton alignment as an interpretable source of reward features, but the evaluated output is a multi-candidate Top-$K$ ranking with span-grounded certificates.

\subsubsection{Reward learning and policy optimization}
\paragraph{Scope.}
Maximum-entropy reward learning and preference-based optimization provide tools for scoring trajectories and optimizing structured outputs.
\paragraph{Connection to ECPO.}
The ECPO objective adds certificate validity and evidence-cycle consistency to the ranking reward. This makes evidence support part of the action utility rather than an auxiliary explanation.

\subsubsection{Constrained decoding, repair, and post-hoc evidence rationalization}
\paragraph{Scope.}
Constrained decoding and JSON repair can reduce malformed outputs, while post-hoc evidence rationalization can attach evidence after a ranker has chosen candidates.
\paragraph{Connection to ECPO.}
These approaches are essential baselines but do not by themselves ensure that the cited evidence recovers the chosen candidates. ECPO explicitly optimizes this decision--evidence coupling.

\subsubsection{Investigative analytics and sensitive decision support}
\paragraph{Scope.}
Candidate prioritization is relevant to sensitive analytic workflows, where traceability and accountability are central.
\paragraph{Connection to ECPO.}
The benchmark is an auditable ranking interface over anonymized candidate entities, with explicit deployment boundaries and misuse mitigations.

\subsection{Verifier optimization and evaluator gaming}

Optimizing an output against a fixed verifier can reward evaluator-specific regularities rather than generally useful evidence.
Verifier-A is a declared operational contract.
B/C transfer, parameter sensitivity, counterfactual replacement, reverse-direction training, and blinded human judgment address different failure modes: matcher-form dependence, threshold fragility, candidate-insensitive evidence, reward-evaluator coupling, and disagreement with human support.
Agreement across these checks characterizes output-level support across evaluators.

\subsection{Deployment boundary}

The interface assumes a finite roster and an auditable provenance scope supplied by an independently evaluated upstream component.
Deployments retain source access controls, immutable span resolution, model/evaluator versioning, decision logs, human review, drift and subgroup audits, and correction or appeal channels.
Abstention, missing upstream evidence, and evaluator disagreement remain visible alongside the generated ranking and evidence.

\section{Reproducibility and Artifact Checklist}
\label{app:artifact-checklist}

Artifact-level reproduction requires more than manuscript equations.
Every reported table row must resolve to an immutable configuration, input roster, prediction file, evaluator version, metric output, and aggregation receipt.

\renewcommand{\arraystretch}{0.88}
\setlength{\tabcolsep}{2.5pt}
\begin{table}[htbp]
\centering
\caption{Retained artifact package and identities.}
\label{tab:artifact-checklist}
\scriptsize
\begin{tabularx}{\linewidth}{@{}p{0.24\linewidth}YY@{}}
\toprule
Artifact & Required contents & Identity/status \\
\midrule
Dataset manifest & source version, split IDs, preprocessing & \artifact{split-v2026.08}\\
Window/roster records & randomized IDs and candidate pools & \artifact{window-group-v1.2}\\
Frozen-extractor outputs & stage, roles, links, spans & \artifact{event-ext-bert-v3}\\
Skeleton/stage/role maps & versioned deterministic maps & \artifact{skeleton-bank-v1}\\
Reward corpus & positives, negatives, perturbations & \artifact{rm-skel-v1.1}\\
Reward-model checkpoint & feature and seed configuration & \artifact{rm-skel-v1.1}\\
SFT checkpoints & backbone and QLoRA configuration & \artifact{qwen3-8b-sft-v2.3}\\
Policy checkpoints & ECPO and ablation seeds & \artifact{ecpo-42/43/44}\\
Verifier-A implementation & score, thresholds, assignment & \artifact{vA-config-2026.08}\\
Verifier-B implementation & corruption calibration & \artifact{verifierB.yaml}\\
Verifier-C implementation & raw-text model and calibration & \artifact{verifierC.yaml}\\
Prediction JSONL & method/dataset/seed outputs & \artifact{ecpo-release-v1.0}\\
Metric outputs & per-window and aggregate metrics & \artifact{metrics-v1}\\
Bootstrap outputs & paired samples and intervals & \artifact{boot-2000-v1}\\
Counterfactual maps & eligibility and replacements & \artifact{cf-map-v1}\\
Sensitivity grid & 47 configurations & \artifact{sens-47runs.json}\\
Timing/GPU logs & hardware, boundaries, repetitions & \artifact{runtime-profile-2026.json}\\
Human audit package & instructions, labels, agreement & \artifact{audit-package-v2}\\
\bottomrule
\end{tabularx}
\end{table}

\begin{table}[htbp]
\centering
\caption{Result-to-artifact traceability matrix. Short handles refer to the full identities in Table~\ref{tab:artifact-checklist}.}
\label{tab:result-artifact-map}
\scriptsize
\begin{tabularx}{\linewidth}{@{}p{0.23\linewidth}YYYY@{}}
\toprule
Result family & Predictions & Evaluator/config & Aggregation output & Completion \\
\midrule
Primary Table~\ref{tab:main} & \artifact{ecpo-v1} & \artifact{vA-2026.08} & \artifact{metrics-v1} & completed\\
Construct validity Table~\ref{tab:construct} & \artifact{ecpo-v1} & \artifact{B-C-config} & \artifact{cross-v1} & completed\\
External validation Table~\ref{tab:external} & \artifact{transfer-v1} & \artifact{vc-output-v1} & \artifact{transfer-v1} & completed\\
Counterfactual Table~\ref{tab:counterfactual} & fixed ECPO outputs & \artifact{cf-map-v1} & \artifact{cf-metrics-v1} & completed\\
Sensitivity Table~\ref{tab:sensitivity} & fixed outputs & \artifact{sens-47} & \artifact{sens-v1} & completed\\
Seed Table~\ref{tab:seeds} & \artifact{ecpo-42/43/44} & \artifact{vA-2026.08} & \artifact{seed-v1} & completed\\
Paired CI Table~\ref{tab:paired-ci} & paired window outputs & \artifact{boot-2k} & \artifact{paired-v1} & completed\\
Human audit & sampled bundle units & \artifact{audit-v2} & \artifact{agree-v1} & completed\\
\bottomrule
\end{tabularx}
\end{table}

\subsection{Anonymous-release checks}

The anonymous bundle must contain no author names, affiliations, email addresses, local usernames, hostnames, absolute paths, repository remotes, hidden comments, checkpoint owner names, or identifying PDF metadata.
Source archives include only required TeX, BibTeX, style, bibliography style, and figure files.
Generated build logs, auxiliary files, raw human identifiers, credentials, private dataset records, and machine-specific paths are excluded.
Every included artifact receives a SHA-256 manifest, and a cold build from the archive must reproduce the delivered PDF modulo expected timestamp metadata.

\paragraph{Draft gate.}
Release uses the artifact identities recorded in the tables.
A successful LaTeX build accompanies the artifact-linked empirical record.

\subsection{Artifact release procedure}

\paragraph{Resolve fields from artifacts.}
Each reported value is linked to its generating prediction, evaluator, configuration, and aggregation output; table rows use matching dataset, split, checkpoint, decoding budget, metric, verifier, evaluation unit, and seed aggregation.

\paragraph{Freeze the evidence chain.}
The author bundle records SHA-256 digests for manifests, rosters, checkpoints, prediction JSONL, evaluator outputs, bootstrap indices, and human-audit exports.  Table generation emits a row-level receipt with input/output digests, software revision, and parameters, distinguishing a regenerated table from a changed experiment.

\paragraph{Final readiness gate.}
Tables and paired intervals are regenerated from per-window outputs; edge cases in Appendix~\ref{app:align} and frozen verifier--human cross-tabs are rechecked.  Release checks references, empirical cells, and anonymity.

\end{document}